# Multivariate Probabilistic CRPS Learning with an Application to Day-Ahead Electricity Prices

Jonathan Berrisch[a,*], Florian Ziel[a]

[a] *Chair of Environmental Economics, esp. Economics of Renewable Energy*
*University of Duisburg-Essen*
*Germany*

**Abstract**

This paper presents a new method for combining (or aggregating or ensembling) multivariate probabilistic forecasts, considering dependencies between quantiles and marginals through a smoothing procedure that allows for online learning. We discuss two smoothing methods: dimensionality reduction using Basis matrices and penalized smoothing. The new online learning algorithm generalizes the standard CRPS learning framework into multivariate dimensions. It is based on Bernstein Online Aggregation (BOA) and yields optimal asymptotic learning properties. The procedure uses horizontal aggregation, i.e., aggregation across quantiles. We provide an in-depth discussion on possible extensions of the algorithm and several nested cases related to the existing literature on online forecast combination. We apply the proposed methodology to forecasting day-ahead electricity prices, which are 24-dimensional distributional forecasts. The proposed method yields significant improvements over uniform combination in terms of continuous ranked probability score (CRPS). We discuss the temporal evolution of the weights and hyperparameters and present the results of reduced versions of the preferred model. A fast C++ implementation of the proposed algorithm is provided in the open-source R-Package *profoc* on CRAN.

*Corresponding author

*Email addresses:* jonathan.berrisch@uni-due.de (Jonathan Berrisch), florian.ziel@uni-due.de (Florian Ziel)

## 1. Introduction

Forecast combination (sometimes referred to as expert aggregation or ensembling) has recently gained much traction. We know from theory that combination methods work well to combine different but well-performing model classes Cesa-Bianchi & Lugosi (2006). As Gaillard et al. (2016) pointed out, it is always recommended to use different classes of models, e.g., regression and time series type models, neural network models, decision tree learning models, and other machine learning and artificial intelligence methods.

This paper proposes a novel online updating scheme for combining the marginals of the corresponding multivariate distribution across quantiles (also referred to as horizontal aggregation). We know from Sklar's theorem that we can decompose any multivariate distribution into the marginals and a copula. That is, we can improve the marginals (i.e., by using a strictly proper scoring rule like the CRPS) while leaving the copula untouched. In consequence, we require only the reporting of the forecasted marginal distribution. The proposed method considers dependencies between the combination weights across quantiles and marginals through a simple but flexible smoothing procedure. We assume a basic metric or spatial structure in the multivariate dimension. Such a metric structure is present when forecasting a univariate time series several steps ahead or predicting one-dimensional spatial data.

Online learning algorithms are particularly attractive for forecasting where frequent short-term forecasts are essential for the application domain (e.g., energy, weather, finance, retail). The proposed algorithm generalizes the probabilistic CRPS learning framework presented in Berrisch & Ziel (2021). It is based on exponential weighted averaging (EWA) and yields optimal asymptotic convergence rates with respect to the best individual forecast and the best convex combination of all forecasts (Wintenberger, 2017).

Considerable research on forecasting combination already exists. Bordignon et al. (2013); Nowotarski et al. (2014); Avci et al. (2018) combine point-forecasts using various batch methods. Marcjasz et al. (2020); Serafin et al. (2019) apply batch methods to probabilistic forecast combination. Some authors also applied online learning algorithms for point forecasting (Nowotarski et al., 2016) and probabilistic forecasting (Gaillard et al., 2016; Gonzalez et al., 2021a). The work above focuses on developing distinct forecasting models and on combination methods. Gaillard & Goude (2015) discuss how model development can be optimized in the framework of aggregation of experts.

In electricity price forecasting, dynamic aggregation techniques, where the combination weights are adjusted based on past performance, tend to perform better than simple constant weight techniques (Gaillard & Goude, 2015; Marcjasz et al., 2018; Maciejowska et al., 2020). However, they consider multivariate updating schemes that use the same weight for all time series. Most other work in energy forecasting considers all time series to be independent and therefore combines forecasts separately (Bordignon et al., 2013; Nowotarski et al., 2016; Nitka & Weron, 2023). Neither approach considers possible dependencies of combination weights between marginals. Consequently, we can expect potential improvements by exploiting this metric structure of electricity prices by considering updating schemes that assign different weights to all neighboring price forecasts of the day and considering possible dependencies between combination weights. Of course, the same logic applies to other areas of application.

The contributions of this manuscript are manifold:

i) We generalize batch and online CRPS learning to multivariate settings.

ii) We show how the metric or spatial structure of the combination weights for multivariate data can be considered using two smoothing methods.

iii) We discuss three possible strategies for optimizing hyperparameters in online learning settings.

iv) We provide a fast C++ implementation of the proposed algorithm in the open-source R-Package *profoc* on CRAN (Berrisch & Ziel, 2023).

v) We empirically apply the proposed methods to multivariate probabilistic day-ahead electricity price forecasts.

The remainder of this paper is structured as follows. Section 2 discusses the general multivariate probabilistic combination setting and discusses CRPS learning using quantile regression. Section 3 presents the proposed multivariate generalization of online CRPS learning and summarizes its asymptotic properties. Additionally, we discuss possible extensions of the proposed method. Those extensions to the core algorithm add hyperparameters that have to be specified. Therefore, we elaborate on two possible strategies for hyperparameter tuning in Section 4. Section 5 continues with an empirical application of the proposed algorithm. We apply the methodology to multivariate probabilistic forecasts of Day-Ahead power prices. We discuss the data, elaborate on the specific algorithms we consider, and present a detailed analysis of the obtained results. Section 6 discusses limitations, introduces potential enhancements, and concludes.

## 2. Multivariate CRPS Learning

### *2.1. The combination setting*

In this paper, we consider the combination of multivariate probabilistic forecasts. In particular, we consider a setting where the forecasts are given as quantiles of all marginals of a multivariate distribution. Berrisch & Ziel (2021) show that pointwise forecast combinations potentially outperform standard methods where weights are constant over all distribution quantiles. We apply this idea to a multivariate setting by computing weights depending on the quantile and the marginals. First, we discuss batch learning methods and propose a dimension reduction technique that bridges the gap between flexible pointwise and robust constant procedures. Afterward, we show how the proposed online learning algorithm of Berrisch & Ziel (2021) can be extended for combining the marginals of multivariate probabilistic forecasts.

Let $\widehat{\boldsymbol{F}}_t = (\widehat{F}_{t,1}, \dots, \widehat{F}_{t,K})$ be a vector of $K$ univariate distributions representing the marginal distribution of the corresponding multivariate distribution, resp. the set of experts that we want to combine. We consider the combination across quantiles (also known as horizontal aggregation):

$$\widetilde{F}_t^{-1} = \sum_{k=1}^{K} w_{t,k} \widehat{F}_{t,k}^{-1} \tag{1}$$

We evaluate the performance using the cumulative CRPS over all marginals. Therefore, the weights shall be chosen to minimize the cumulative CRPS of all marginals. We can approximate the CRPS by the sum over Quantile Losses (QL)

$$\text{CRPS}(F, y) = \int_{\mathbb{R}} (F(x) - \mathbb{1}\{x > y\})^2 dx \approx \frac{2}{P} \sum_{p \in \boldsymbol{\mathcal{P}}} \text{QL}_p(F^{-1}(p), y) \tag{2}$$

for an equidistant dense grid $\boldsymbol{\mathcal{P}} = (p_1, \dots, p_P)$ with $p_i < p_{i+1}$ and $p_{i+1} - p_i = h$ for all $p$. Clearly, $P \to \infty$ induces $h \to 0$, $p_1 \to 0$, $p_P \to 1$ and the approximation converges to the CRPS (Gneiting, 2011a,b). Marcjasz et al. (2023) omitted the scaling factor of 2 in equation (2) as it does not affect the optimization, and there is no natural interpretation of the CRPS. We follow this approach to ensure comparability of the results.

This relationship enables us to compute pointwise weights based on quantile losses. We can extend this idea by optimizing weights not only depending on the quantile $p$ but also on the marginal $d$:

$$\widetilde{F}_{t,d}^{-1}(p) = \sum_{k=1}^{K} w_{t,k}(d,p)\widehat{F}_{t,k,d}^{-1}(p) \tag{3}$$

We are interested in setting $w_{t,k}$ such that the CRPS of $\widetilde{F}_{t,d}$ is minimized.

### *2.2. CRPS learning using quantile regression*

Pointwise CRPS learning has the potential to outperform standard CRPS learning methods. However, the best pointwise weights in (3) must be estimated. Theoretically, a pointwise approach has to be applied to all probabilities $p \in (0,1)$ and all marginals $\boldsymbol{\mathcal{D}} = (1, 2, \dots, D)$ such that the bivariate weight function $\boldsymbol{w}_{t,k}$ can be specified. However, we can never evaluate infinitely many values for $p$. On the same page, the computation may be infeasible if $D$ is very large. Therefore, we must consider some finite-dimensional representation for the weight functions $\boldsymbol{w}_{t,k}$. A suitable option is representing the weight functions $\boldsymbol{w}_{t,k}$ using a finite-dimensional representation using splines. Bivariate splines are a suitable option in this scenario. We can express them as follows:

$$f(X_{t,j_1}, X_{t,j_2}) = \sum_{l=1}^{L} \beta_l \boldsymbol{\varphi}_l(X_{t,j_1}, X_{t,j_2}). \tag{4}$$

This is essentially the same as univariate splines with $L$-dimensional parameter vector $(\beta_1, \dots, \beta_L)'$. However, the support of $f$ is 2-dimensional. Thus, we need many more basis functions $L$ to have a suitable description of $f$. A popular way to describe the bivariate basis function $\boldsymbol{\varphi}_l$ in (4) is to assume a tensor structure (McLean et al., 2014; Wood et al., 2017). In the bivariate case, the spline function is a product of two univariate ones. In addition, $\varphi_{1,l}$ and $\varphi_{2,l}$ are usually chosen such that $\varphi_{1,l_1}$ interacts with each of the considered basis functions $\varphi_{2,l_2}$. This, allows to renumerate the problem such that $l = (l_1, l_2)$, and yields

$$\boldsymbol{\varphi}_{l_1,l_2}(x_1, x_2) = \varphi_{1,l_1}(x_1)\varphi_{2,l_2}(x_2). \tag{5}$$

This can be used to express the bivariate weight function as a product of the $\widetilde{P} \times \widetilde{D}$ parameter matrix $\boldsymbol{\beta}_{t,k}$ and the bivariate basis represented by $\boldsymbol{\varphi}^{\text{mv}}$ and $\boldsymbol{\varphi}^{\text{pr}}$:

$$\boldsymbol{w}_{t,k} = \sum_{j=1}^{\widetilde{D}} \sum_{l=1}^{\widetilde{P}} \beta_{t,j,l,k} \varphi_j^{\text{mv}} \varphi_l^{\text{pr}} = \boldsymbol{\varphi}^{\text{mv}} \boldsymbol{\beta}_{t,k} \boldsymbol{\varphi}^{\text{pr}\prime}. \tag{6}$$

Given $T$ historic forecasts $\widehat{F}^{-1}_{t,d,k}(p)$, the cooresponding realizations $Y_{t,d}$, and the index of marginals $\boldsymbol{\mathcal{D}} = (1, 2, \dots, D)$ we can estimate the $\widetilde{D} \times \widetilde{P} \times K$-dimensional parameter tensor $\boldsymbol{\beta}_t$ by minimizing the corresponding CRPS using (2):

$$\boldsymbol{\beta}^{\boldsymbol{\varphi}\text{-CRPS}}_t = \operatorname*{arg\,min}_{\boldsymbol{\beta} \in \mathbb{R}^{K \times L}} \sum_{i=t-T+1}^{t} \sum_{d \in \boldsymbol{\mathcal{D}}} \int_0^1 \rho_p \left( Y_{t,d} - \sum_{k=1}^{K} \sum_{j=1}^{\widetilde{D}} \sum_{l=1}^{\widetilde{P}} \beta_{t,j,l,k} \varphi^{\text{mv}}_{d,j} \varphi^{\text{pr}}_{l,p} \widehat{F}^{-1}_{t,d,k}(p) \right) dp. \tag{7}$$

The second line uses the shift-invariance of the quantile loss and quantile regression notation $\rho_p(z) = \text{QL}_p(0, z) = z(p - \mathbb{1}\{z < 0\})$ (Koenker, 2017).

Still, computing (7) requires the evaluation of all distribution forecasts. As discussed, this is often not possible in practice. If we restrict the evaluation to a grid of probabilities $\boldsymbol{\mathcal{P}}$ problem (7) simplifies with (2) to

$$\boldsymbol{\beta}^{\boldsymbol{\varphi}\text{-QR}}_t = \operatorname*{arg\,min}_{\boldsymbol{\beta} \in \mathbb{R}^{K \times L}} \sum_{i=t-T+1}^{t} \sum_{d \in \boldsymbol{\mathcal{D}}} \sum_{p \in \boldsymbol{\mathcal{P}}} \rho_p \left( Y_{t,d} - \sum_{k=1}^{K} \sum_{j=1}^{\widetilde{D}} \sum_{l=1}^{\widetilde{P}} \beta_{t,j,l,k} \varphi^{\text{mv}}_{d,j} \varphi^{\text{pr}}_{l,p} \widehat{F}^{-1}_{t,d,k}(p) \right). \tag{8}$$

In general, quantile regression problems can be solved efficiently using linear programming solvers (Koenker, 2017). However, (8) is not a simple quantile regression problem, but a joint quantile regression (Sangnier et al., 2016; Chun et al., 2016). The parameters $\beta_{t,j,l,k}$ are active for multiple quantiles. Thus, adequate estimation requires solving the optimization problem for $\widetilde{D} \times \widetilde{P} \times K$ parameters, which can be computationally costly if $\widetilde{D}$, $\widetilde{P}$, and $K$ are large.

However, if we choose both basis $\boldsymbol{\varphi}^{\text{pr}}$ and $\boldsymbol{\varphi}^{\text{mv}}$ so that $\boldsymbol{\varphi}^{\text{mv}}_i = \mathbb{1}\{d_i\}$ on $\boldsymbol{\mathcal{D}}$ for $d_i \in \boldsymbol{\mathcal{D}} = (d_1, \dots, d_D)$ and $\boldsymbol{\varphi}^{\text{pr}}_i = \mathbb{1}\{p_i\}$ on $\boldsymbol{\mathcal{P}}$ for $p_i \in \boldsymbol{\mathcal{P}} = (p_1, \dots, p_P)$ then (8) can be disentangled into $\widetilde{D} \times \widetilde{P}$ separate quantile regression problems. This is

$$\boldsymbol{w}^{\text{QR}}_{t,h}(p) = \operatorname*{arg\,min}_{\boldsymbol{w} \in \mathbb{R}^{K}} \sum_{i=t-T+1}^{t} \rho_p \left( Y_{i,d} - \sum_{k=1}^{K} w_{i,d,k} \widehat{F}^{-1}_{i,d,k}(p) \right) \tag{9}$$

for $p \in \boldsymbol{\mathcal{P}}$ and $d \in \boldsymbol{\mathcal{D}}$ where $\widehat{F}^{-1}_{t,h,k}(p)$ are the experts for the $p$-quantile and the $d$-marginal.

Quantile regression (9) will lead to linear optimality on $\boldsymbol{\mathcal{D}}$ and $\boldsymbol{\mathcal{P}}$, as long as standard regularity conditions required for the quantile regression are satisfied (Koenker & Hallock, 2001). However, we might assume further restrictions to reduce the estimation risk, e.g., the solution is a convex combination. Taylor & Bunn (1998) discussed many related plausible restrictions for quantile combination concerning bias correction, positivity, and affinity, among others.

A potential issue of pointwise algorithms is *quantile crossing*. This problem occurs if we have $\widetilde{F}_{t,d}^{-1}(p_i) > \widetilde{F}_{t,h}^{-1}(p_j)$ for some $p_i, p_j \in (0,1)$ with $p_i < p_j$. In this case, we recommend rearranging the predictions as sorting is known to improve the forecasting performance (Chernozhukov et al., 2010).

## 3. Multivariate Online CRPS Learning

Batch-learning approaches, like quantile regression, evaluate the entire history for estimating new combination weights, which is computationally costly. Therefore, we suggest to use online learning methods instead.

Online learning is often called *prediction under expert advice*. In this context, *experts* refer to the models producing the predictions (or predictive distributions). The person or model that combines the experts' predictions is called *forecaster*. A key element of online learning methods is (cumulative) regret. It is defined as:

$$R_{t,k} = \sum_{i=1}^{t} r_{t,k} = \sum_{i=1}^{t} \ell(\widetilde{X}_i, Y_i) - \ell(\widehat{X}_{i,k}, Y_i) \tag{10}$$

i.e., the cumulative difference between the loss of the expert's predictions $\widehat{X}_{t,k}$ and the prediction of the forecaster $\widetilde{X}_t$ for a loss function $\ell$. $\widetilde{X}_{t,k}$ might be a predicted quantile $\widehat{F}_{t,k}^{-1}(p)$ of expert $k$ as discussed in the previous section. $R_{t,k}$ is called regret because it indicates how much the forecaster regrets not following the experts' advice (Cesa-Bianchi & Lugosi, 2006). With (10), we can formulate the EWA:

$$w_{t,k}^{\text{EWA}} = K w_{0,k} \frac{e^{\eta R_{t,k}}}{\sum_{j=1}^{K} e^{\eta R_{t,j}}} = \frac{e^{-\eta \ell(\widehat{X}_{t,k}, Y_t)} w_{t-1,k}^{\text{EWA}}}{\sum_{j=1}^{K} e^{-\eta \ell(\widehat{X}_{t,j}, Y_t)} w_{t-1,j}^{\text{EWA}}} \tag{11}$$

where $K$ refers to the number of experts, $w_{0,k}$ to the initial weights of an expert $k$, and $\eta$ to the learning rate, which defines how fast the weights adjust to changes in the regret Cesa-Bianchi & Lugosi (2006). We can express this aggregation rule in terms of past weights and the loss suffered by the experts (right-hand side of 11). This highlights that there is no need for evaluating the entire history when adjusting weights.

EWA yields optimal convergence rates of $\mathcal{O}(T)$ towards the best expert for exp-concave loss functions Cesa-Bianchi & Lugosi (2006). It means that the algorithm's performance (in terms of risk) is asymptotically not worse than the performance of the best expert. A more ambitious property that can also be satisfied is the *convex aggregation property*. It ensures that the risk of the algorithm is not worse than the risk of the best convex combination of the experts. For an algorithm to satisfy this property, the *gradient trick* is needed (Devaine et al., 2013). For exp-concave losses, this gives optimal convergence rate $\mathcal{O}(\sqrt{T})$ with respect to the best convex combination of the experts (Cesa-Bianchi & Lugosi, 2006). This property also holds for losses that satisfy some Bernstein condition, such as the MAE, when algorithms like Bernstein Online Aggregation (BOA) are used. These algorithms use regularized updating techniques to improve convergence and stability properties (Wintenberger, 2017).

Berrisch & Ziel (2021) adapted BOA to probabilistic problems. The new algorithm is called CRPS learning because it optimizes the CRPS of the target distribution using pointwise optimization on a grid of quantiles. The weights can vary over time and in different parts of the distribution. CRPS learning still maintains the fast convergence of BOA. This algorithm for combining $\widehat{\boldsymbol{X}}_t = (\widehat{X}_{t,1}, \dots, \widehat{X}_{t,K})$ to $\widetilde{X}_t = \boldsymbol{w}'_{t-1}\widehat{\boldsymbol{X}}_t$ can be summarized as follows:

$$\boldsymbol{r}_t = \text{QL}^{\nabla}_{\boldsymbol{\mathcal{P}}}(\widetilde{X}_t, Y_t) - \text{QL}^{\nabla}_{\boldsymbol{\mathcal{P}}}(\widehat{\boldsymbol{X}}_t, Y_t) \tag{12a}$$

$$\boldsymbol{E}_t = \max(\boldsymbol{E}_{t-1}, \boldsymbol{r}_t^{+} + \boldsymbol{r}_t^{-}) \tag{12b}$$

$$\boldsymbol{V}_t = \boldsymbol{V}_{t-1} + \boldsymbol{r}_t^{\odot 2} \tag{12c}$$

$$\boldsymbol{\eta}_t = \min\left(\left(-\log(\boldsymbol{w}_0) \odot \boldsymbol{V}_t^{\odot -1}\right)^{\odot \frac{1}{2}}, \frac{1}{2}\boldsymbol{E}_t^{\odot -1}\right) \tag{12d}$$

$$\boldsymbol{R}_t = \boldsymbol{R}_{t-1} + \boldsymbol{r}_t \odot (\boldsymbol{1} - \boldsymbol{\eta}_t \odot \boldsymbol{r}_t)/2 + \boldsymbol{E}_t \odot \mathbb{1}\{-2\boldsymbol{\eta}_t \odot \boldsymbol{r}_t > 1\} \tag{12e}$$

$$\boldsymbol{w}_t = K\boldsymbol{w}_0 \odot \text{SoftMax}\left(-\boldsymbol{\eta}_t \odot \boldsymbol{R}_t + \log(\boldsymbol{\eta}_t)\right) \tag{12f}$$

where $\boldsymbol{x}^+$ and $\boldsymbol{x}^-$ denote the elementwise positive and negative parts of $\boldsymbol{x}$ and $\odot$ the elementwise product (Hadamard product). The learning rate $\boldsymbol{\eta}_t$ determines the weight adjustment speed. It depends on the bound estimator $\boldsymbol{E}_t$ and $\boldsymbol{V}_t$, which is an estimator for the variance. The algorithm describes how weights are calculated on a full quantile grid $\boldsymbol{\mathcal{P}}$. First, the instantaneous regret is calculated (12a). Then the learning rate (12b-12d) is adjusted. In (12e) the cumulative regret is calculated. Afterward, we calculate the weights (12f). Finally, the forecaster uses $\boldsymbol{w}_t$ to calculate $\widetilde{X}_{t+1}$ and starts over with (12a).

Several extensions of online learning algorithms were proposed in the literature. However, they can also be applied in standard Batch-Learning settings. The benefits of these extensions have been confirmed in empirical studies. Some extensions, like shrinkage operators, are also valuable to nest specific weighting strategies into the learning algorithm.

### 3.1. Smoothing

As mentioned, we apply the general CRPS learning idea to multivariate data. Therefore, we adopt the two weight-smoothing methods of the original CRPS Learning algorithm. The first consists of the dimension reduction method using basis matrices. The approach is analogous to the idea discussed in Section 2.2. Using a bivariate basis, we can reduce the dimensionality of the instantaneous regret from $D \times P$ to $\widetilde{D} \times \widetilde{P}$:

$$\widetilde{\boldsymbol{r}}_k = \frac{\widetilde{D}\widetilde{P}}{DP} \boldsymbol{B}^{\text{mv}\prime} \boldsymbol{r}_k \boldsymbol{B}^{\text{pr}}. \tag{13}$$

As usual, we can use this reduced regret to carry out the online learning algorithm. After obtaining weights (we refer to them as $\boldsymbol{\beta}_{t,k}$) on this condensed version of the regret, we can utilize the basis matrices once again to obtain weights in our original dimensions of interest $\boldsymbol{w}_{t,k} = \boldsymbol{B}^{\text{mv}} \boldsymbol{\beta}_{t,k} \boldsymbol{B}^{\text{pr}\prime}$.

This relatively simple method yields a powerful property: It bridges the gap between purely pointwise weight optimization based on quantiles and the constant approach where a single weight is optimized with respect to the CRPS. That means we can move from a setting with low flexibility (i.e., a few parameters) and low estimation risk to a very flexible one (with many parameters) at the price of high estimation risk.

Another option is to smooth the weights using penalized smoothing. This method can be applied after the estimation, i.e., after the updating step. Hereby we consider two sets of bounded basis functions $\boldsymbol{\psi}^{\text{mv}} = (\psi_1, \dots, \psi_D)$ and $\boldsymbol{\psi}^{\text{pr}} = (\psi_1, \dots, \psi_P)$ on $(0, 1)$ that we will use for penalized smoothing.

Then the weights can be represented by

$$\boldsymbol{w}_{t,k} = \boldsymbol{\psi}^{\text{mv}} \boldsymbol{b}_{t,k} \boldsymbol{\psi}^{\text{pr}\prime} \tag{14}$$

with parameter matix $\boldsymbol{b}_{t,k}$. We estimate $\boldsymbol{b}_{t,k}$ by penalized $L_1$- and $L_2$-smoothing which minimizes

$$\begin{aligned} &\|\boldsymbol{\beta}'_{t,d,k} \boldsymbol{\varphi}^{\text{pr}} - \boldsymbol{b}'_{t,d,k} \boldsymbol{\psi}^{\text{pr}}\|_2^2 + \lambda^{\text{pr}} \|\mathcal{D}_q (\boldsymbol{b}'_{t,d,k} \boldsymbol{\psi}^{\text{pr}})\|_2^2 + \\ &\|\boldsymbol{\beta}'_{t,p,k} \boldsymbol{\varphi}^{\text{mv}} - \boldsymbol{b}'_{t,p,k} \boldsymbol{\psi}^{\text{mv}}\|_2^2 + \lambda^{\text{mv}} \|\mathcal{D}_q (\boldsymbol{b}'_{t,p,k} \boldsymbol{\psi}^{\text{mv}})\|_2^2 \end{aligned} \tag{15}$$

for each $k$ given $\boldsymbol{\beta}_{t,k}$ with differential operator $\mathcal{D}_q$ of order $q$. The differential order characterizes the smoothing penalty, and $\lambda \geq 0$ characterizes the roughness penalty.

Typically, $q = 2$ is considered along with cubic B-Splines to penalize for roughness (Wang, 2011; Wood, 2017). However, we prefer using $q < 2$ here. This smoothes towards constant weights over $\boldsymbol{\mathcal{P}}$ for $\lambda \to \infty$ and not towards a linear relationship between weights and probabilities as for $q = 2$. As pointed out in Berrisch & Ziel (2021) no argument supports shrinkage towards a linear relationship. In contrast, shrinkage towards constant weights yields the non-pointwise CRPS-learning theory of constant weight functions. However, let us remark that the penalized smoothing approach with $\lambda \to \infty$ yields a different result than the simple basis smoothing approach mentioned before with $\boldsymbol{\varphi} = \varphi_1 \equiv 1$.

In applications, we only apply this function bases approach on finite grids of probabilities $\boldsymbol{\mathcal{P}} = (p_1, \dots, p_P)$ and a finite number of marginals $\boldsymbol{\mathcal{D}} = (1, \dots, D)$. If we consider B-Spline basis functions $\boldsymbol{\psi}^{\text{mv}}$ and $\boldsymbol{\psi}^{\text{pr}}$, then an explicit solution based on ordinary least squares exists for (15). This explicit solution has a ridge regression representation. The smoothed weights matrix $\boldsymbol{w}_{t,k}$ is then given by

$$
\begin{aligned}
\boldsymbol{w}_{t,k}(\boldsymbol{\mathcal{P}}) = & \left(\boldsymbol{B}^{\text{mv}} \left(\boldsymbol{B}^{\text{mv}\prime}\boldsymbol{B}^{\text{mv}} + \lambda \boldsymbol{D}_q^{\text{mv}\prime}\boldsymbol{D}_q^{\text{mv}}\right)^{-1} \boldsymbol{B}^{\text{mv}\prime}\right) \\
& \boldsymbol{\varphi}^{\text{mv}}\left(\boldsymbol{\mathcal{D}}\right) \boldsymbol{\beta}_{t,k} \boldsymbol{\varphi}^{\text{pr}}\left(\boldsymbol{\mathcal{P}}\right)^{\prime} \\
& \left(\boldsymbol{B}^{\text{pr}} \left(\boldsymbol{B}^{\text{pr}\prime}\boldsymbol{B}^{\text{pr}} + \lambda \boldsymbol{D}_q^{\text{pr}\prime}\boldsymbol{D}_q^{\text{pr}}\right)^{-1} \boldsymbol{B}^{\text{pr}\prime}\right) \\
= & \boldsymbol{\mathcal{H}}^{\text{mv}} \boldsymbol{\varphi}^{\text{mv}} \boldsymbol{\beta}_{t,k} \boldsymbol{\varphi}^{\text{pr}\prime} \boldsymbol{\mathcal{H}}^{\text{pr}}
\end{aligned} \tag{16}
$$

with basis matrices $\boldsymbol{B}^{\text{mv}} = \boldsymbol{\psi}^{\text{mv}}\left(\boldsymbol{\mathcal{D}}\right)$ and $\boldsymbol{B}^{\text{pr}} = \boldsymbol{\psi}^{\text{pr}}\left(\boldsymbol{\mathcal{P}}\right)$, penalty matrices $\boldsymbol{D}_q^{\text{mv}}$ and $\boldsymbol{D}_q^{\text{pr}}$. We can easily compute the penalty matrix if the b-spline basis has equidistant knots. Hereinafter, we distinguish $\boldsymbol{D}_q^S$ and $\boldsymbol{D}_q^G$, which refer to the equidistant case and the general case where knot placement does not have to be equidistant, respectively. Let $\Delta$ denote the matrix difference operator:

$$
\boldsymbol{\Delta} = \begin{bmatrix} -1 & 1 & & & \\ & -1 & 1 & & \\ & & \ddots & \ddots & \\ & & & -1 & 1 \end{bmatrix}. \tag{17}
$$

Now, $\boldsymbol{D}_q^S$ can be easily computed as $\boldsymbol{D}_q^S = \Delta^q \boldsymbol{I}$. The computation of $\boldsymbol{D}_q^G$ is more intricate since non-equidistant knots are permitted. The calculation involves an additional weighting step with respect to the non-equidistant distribution of the knots. We elaborate on this topic briefly since the literature is surprisingly scarce (Li & Cao,

2022, Section 2.2). The required difference matrix can be computed as

$$\boldsymbol{D}_q^G = \boldsymbol{W}_q^{-1} \boldsymbol{\Delta} \boldsymbol{W}_{q-1}^{-1} \boldsymbol{\Delta} \cdots \boldsymbol{W}_1^{-1} \boldsymbol{\Delta} \tag{18}$$

where $W_q$ are weighting matrices that depend on the order of the B-Spline basis, denoted as $o$, and the knots. Let $J$ denote the number of inner knots. The dimension of the difference matrices $\Delta$ in (18) depend on $W_q$ are $(J + o - q) \times (J + o - q + 1)$. The total number of knots will be $J + 2o$. We can specify the weighting matrices as:

$$\boldsymbol{W}_q = \frac{1}{o-q} \begin{bmatrix} t_{o+1} - t_{1+q} & & & \\ & t_{o+2} - t_{2+q} & & \\ & & \ddots & \\ & & & t_{J+2o-q} - t_{J+o} \end{bmatrix}, \tag{19}$$

The quantity $o - q$ represents the lag used to differentiate the knots. If the knots are equidistant, then $\boldsymbol{W}_q$ will be proportional to the identity matrix $\boldsymbol{I}$. Therefore, it nets the standard P-Spline, which uses $\boldsymbol{D}_q^S = \Delta^q \boldsymbol{I}$. This gives rise to the general P-Spline estimator. However, $\boldsymbol{W}_q$ is only proportional to $\boldsymbol{I}$ rather than equal to it. Therefore, scaling needs to be applied for the results of both estimators to coincide. The scaling can be applied to lambda, the penalty matrix, or the difference matrix. To state this formally, let $\boldsymbol{P}_q^S = \boldsymbol{D}_q^{S\prime} \boldsymbol{D}_q^S$ and $\boldsymbol{P}_q^G = \boldsymbol{D}_q^{G\prime} \boldsymbol{D}_q^G$ denote the penalty terms of the standard and general P-Spline estimators. The scaling factor with respect to the penalty $\boldsymbol{P}_q^G$ is $(\mathrm{Tr}(\boldsymbol{W}_q) / (J + o - q))^{2q}$ so the following relation holds:

$$\boldsymbol{P}_q^S = (\mathrm{Tr}(\boldsymbol{W}_q) / (J + o - q))^{2q} \boldsymbol{P}_q^G. \tag{20}$$

This is only valid for equidistant B-Splines. For non-equidistant B-Splines, the generalized P-Spline is the only appropriate estimator. However, $\boldsymbol{P}_q^G$ should always be scaled to ensure the comparability between lambda values in equidistant and non-equidistant situations.

For notational brevity, we denote the first and last part of (16) as $\boldsymbol{\mathcal{H}}^{\mathrm{mv}}$ and $\boldsymbol{\mathcal{H}}^{\mathrm{pr}}$, respectively, the so-called hat matrices. Fortunately, they do not depend on time-varying components; therefore, we can compute them prior to the main online learning task, which yields a great reduction in the algorithm's computational complexity.

### *3.2. Knot placement for Smoothing Splines*

For both smoothing approaches discussed above, the knots of the B-Spline basis must be placed. A well-established approach is placing plenty of equidistant knots.

However, as discussed above, non-equidistant knot placement is valid if the penalty is defined accordingly. We consider equidistant and non-equidistant knots. Thereby, the non-central beta distribution with the following parameterization is used for distributing the knots:

$$\mathcal{B}(x, a, b, c) = \sum_{j=0}^{\infty} e^{-c/2} \frac{\left(\frac{c}{2}\right)^j}{j!} I_x \left(a + j, b\right) \tag{21}$$

Where $I_x$ is the incomplete beta function, $a$ and $b$ are shape parameters, and $c$ is the non-centrality parameter Johnson et al. (1995). Algorithm 1 describes the knot placement in detail. It returns equidistant knots if $\mu = 0.5$, $\sigma = 1$, $c = 0$ and the tailweight parameter $\tau = 1$. Figure 1 shows B-Spline Basis' for different knot

**Algorithm 1:** Knot placement for B-Splines

**Data:** Let $\mathcal{B}$ denote the CDF of the beta distribution and $\mu, \sigma, c, \tau, \deg$ be parameters for adjusting the knot placement.

**Result:** Sequence of knots with length $J + 2\left(\deg + 1\right) = J + 2o$

x $= (0, 1, \dots, J + 2)/(J + 2)$

a $= 2\sigma\left(1 - \mu\right)$

b $= 2\sigma\mu$

knots_c $= \mathcal{B}\left(\mathrm{x}, \mathrm{a}, \mathrm{b}, |c|\right)$

**if** $c < 0$ **then**

  knots_c $=$ reverse $\left(1 - \text{knots_c}\right)$

knots_l $= |\tau| \left(\text{knots_c}[2] - \text{knots_c}[1]\right) \left(-\deg, \dots, -1\right)$

knots_r $= |\tau| \left(\text{knots_c}[J + 2] - \text{knots_c}[J + 1]\right) \left(1, \dots, \deg\right) + 1$

knots $=$ combine (knots_l, knots_c, knots_r)

placements for the inputs of Algorithm 1.

### *3.3. Shrinkage operators and Forgetting*

Shrinkage operators are well-known in statistical learning theory. They help to reduce the overfitting problem by shrinking a solution. The P-Spline smoothing discussed above can also be interpreted as a shrinkage operator. However, simple shrinkage operators can also be applied to $\boldsymbol{\beta}_t$. We consider three additional shrinkage operators: the fixed share operator $\mathcal{F}$, the soft-thresholding operator $\mathcal{S}$, and the

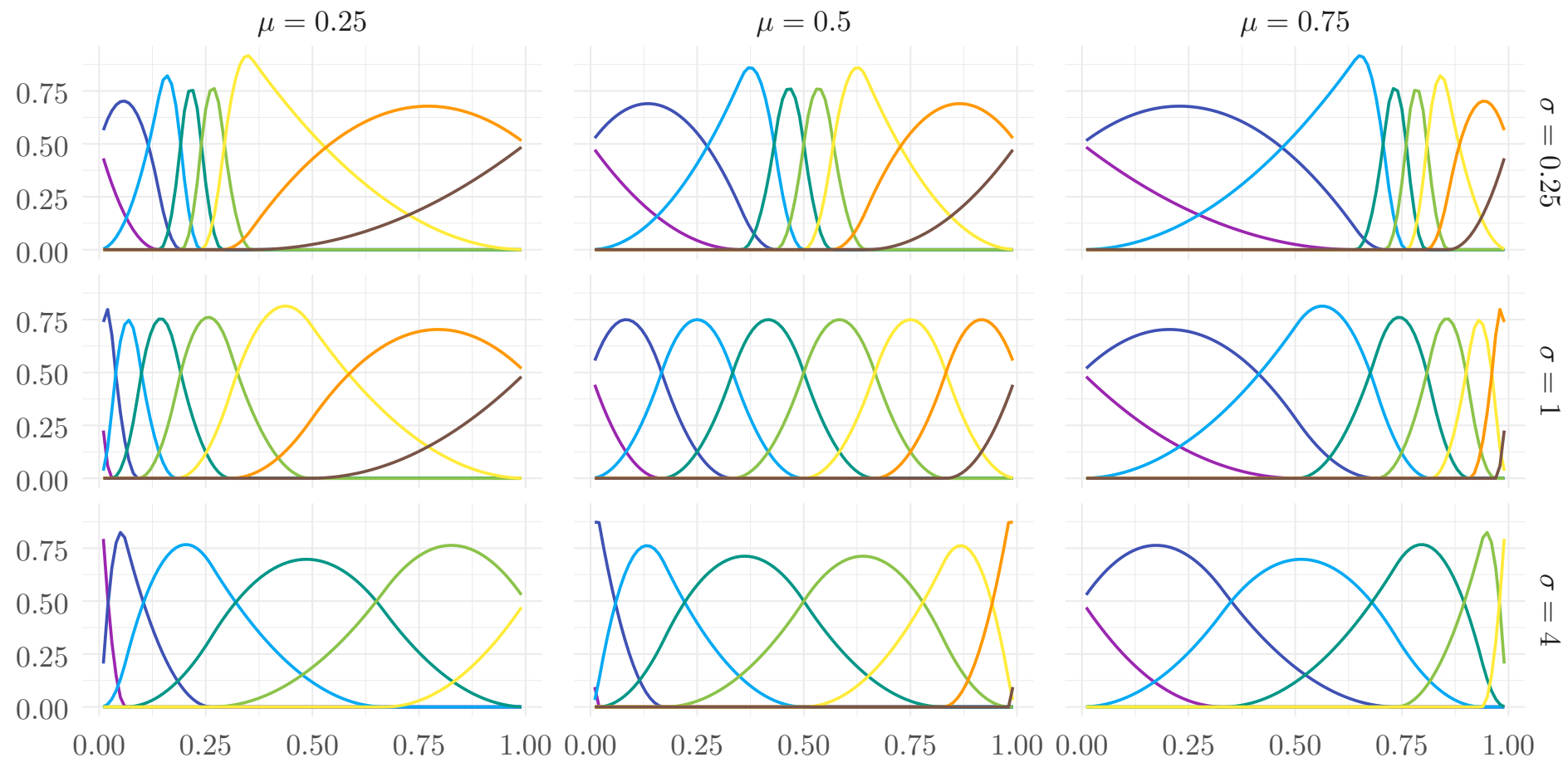

(a) Basis functions for different location and scale values $\mu$ and $\sigma$ (using $\tau = 1$ and $c = 0$)

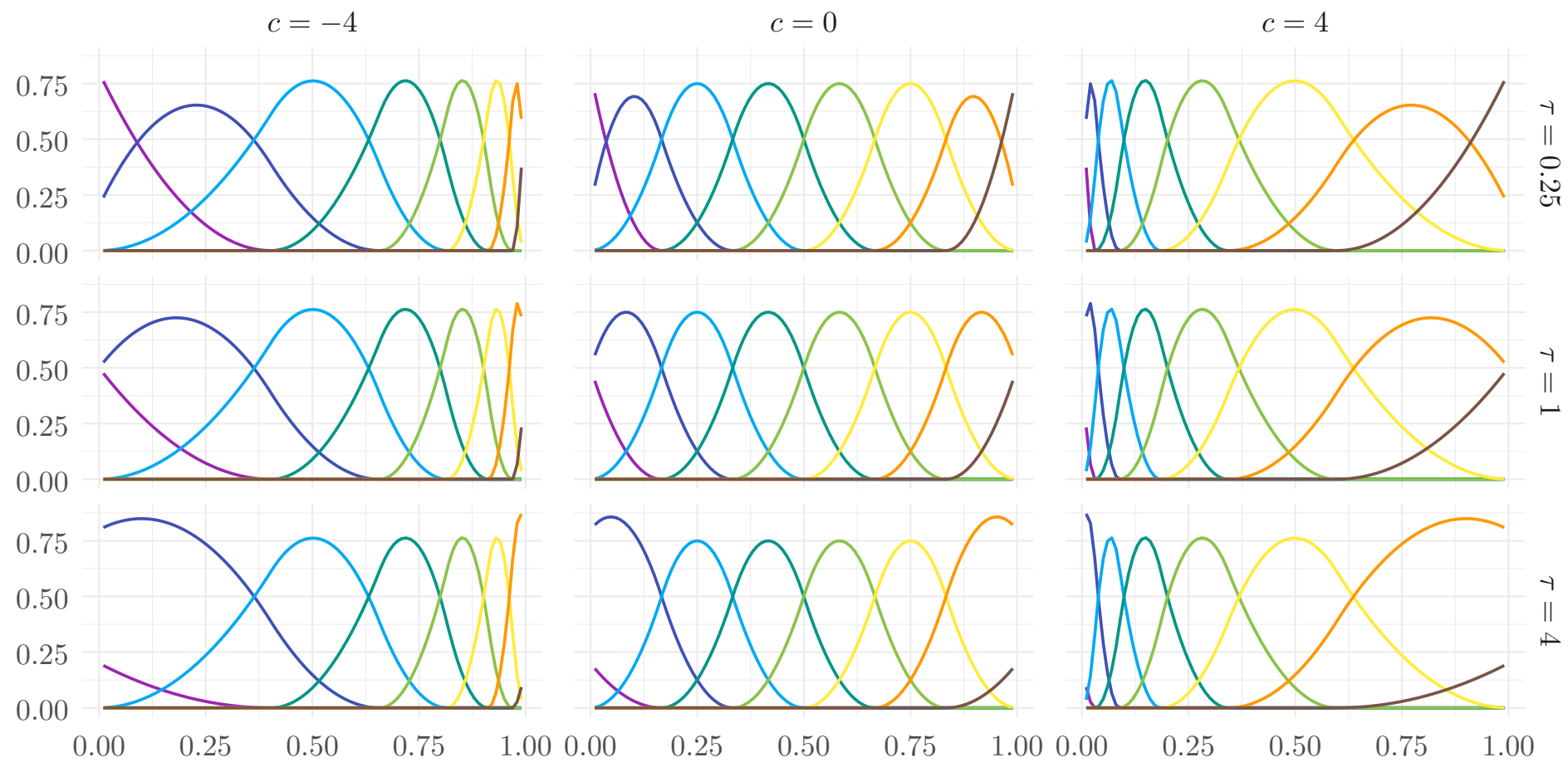

(b) Basis functions for different noncentrality and tailweight values $c$ and $\tau$ (using $\mu = 0.5$ and $\sigma = 1$)

Figure 1: B-Spline functions for selected placements of the knots concerning the inputs of Algorithm 1. The center of both figures shows the default case of equidistant knots.

hard-thresholding operator $\mathcal{H}$. They are defined as

$$\mathcal{F}(x; \phi) = \phi/K + (1 - \phi)x, \tag{22}$$

$$\mathcal{S}(x; \nu) = \text{sign}(x)||x| - \nu|, \tag{23}$$

$$\mathcal{H}(x; \kappa) = x\mathbb{1}\{|x| > \kappa\} \tag{24}$$

for some $\phi \in [0, 1]$, $\nu \geq 0$ and $\kappa \geq 0$. The fixed share operator shrinks towards the naive combination. This is preferable if no prior information on the experts' performance is available. For some shrinkage problems, there are theoretical guarantees for improvements Tu & Zhou (2011); Cesa-Bianchi et al. (2012). Applications in the context of online learning include, e.g., Cesa-Bianchi et al. (2012); Gonzalez et al. (2021b). The thresholding operators $\mathcal{S}$ and $\mathcal{H}$ were also considered in online learning contexts previously (Dalalyan et al., 2012; Gaillard & Wintenberger, 2017). Applying thresholds leads to sparse solutions. Both appear in several situations for specific linear model estimators. Most notably, soft-thresholding is the key operator in the coordinate descent algorithm for estimating the lasso (Friedman et al., 2007). Applying any threshold operator potentially violates affinity constraints (incl. the convexity constraint). Therefore, projections to the desired solution space should be applied.

As mentioned, cumulative regret is a key element in online learning. However, in settings with a long history, there might be structural breaks in the data. These breaks motivate the introduction of the forgetting factor. It means that only a limited amount of the old cumulative regret is considered for adjusting the weights. In other words, the algorithm *forgets* about some part of the past performance. In online learning, usually, exponential forgetting is chosen Guo et al. (2018); Messner & Pinson (2019); Ziel (2022). The Regret with a forgetting factor $\theta \in [0, 1]$ is formally defined as

$$R_{t,k}(\theta) = (1 - \theta)R_{t-1,k} + \ell(\widetilde{F}_t, Y_t) - \ell(\widehat{F}_{t,k}, Y_t) \tag{25}$$

where $\theta = 0$ correspons to no forgetting. Optimal values for the forgetting factor $\theta$ are usually close to 0. The forget should be applied to all hidden state variables in sophisticated online learning procedures like BOA.

## 4. Full Model and Hyperparameter Optimization

Algorithm 2 shows the multivariate online CRPS-Learning algorithm. This includes all extensions discussed in Subsections 3.1 and 3.3. Considering all extensions, this algorithm contains five general hyperparameters (the forget rate $\phi$, the parameters of the shrinkage operators $\theta$, $\kappa$, $\nu$) as well as 30 hyperparameters concerning the design of basis and hat matrices.

The algorithm is versatile as it handles several special cases discussed in the literature. One such case is the uniform combination, also known as the **naive** combination. This can be calculated using the Fixed-Share operator $\mathcal{F}_\phi$ with $\phi = 1$, resulting in uniform weights. There are more efficient ways of calculating uniform weights. However, adjusting the value of $\phi$ allows a smooth transition from the uniform solution to the

**Algorithm 2:** Smoothed CRPS Bernstein Online Aggregation

Initialization see Appendix A

**for** $t$ *in* $1, \dots, T$ **do**

  **for** $d$ *in* $1, \dots, D$ **do** **for** $p$ *in* $1, \dots, P$ **do**

    $\widetilde{X}_{t,d,p} = \boldsymbol{w}'_{t-1,d,p} \widehat{\boldsymbol{X}}_{t,d,p}$

    **for** $k$ *in* $1, \dots, K$ **do** $\boldsymbol{r}^{\text{full}}_{d,p,k} = QL^{\nabla}_{p}(\widetilde{X}_{t,d,p}, Y_t) - QL^{\nabla}_{p}(\widehat{X}_{t,d,p,k}, Y_t)$

  **for** $k$ *in* $1, \dots, K$ **do**

    $\boldsymbol{r}^{\text{red}}_{k} = \frac{\widetilde{D}}{D} \frac{\widetilde{P}}{P} \boldsymbol{B}^{\text{mv}\prime} \boldsymbol{r}^{\text{full}}_{k} \boldsymbol{B}^{\text{pr}}$ // now $\boldsymbol{r}^{\text{red}}_{k}$ is $\widetilde{D} \times \widetilde{P}$

  **for** $d$ *in* $1, \dots, \widetilde{D}$ **do**

    **for** $p$ *in* $1, \dots, \widetilde{P}$ **do**

      $\boldsymbol{V}_{t,d,p} = (1-\theta)\boldsymbol{V}_{t-1,d,p} + \left(\boldsymbol{r}^{\text{red}}_{d,p}\right)^{\odot 2}$

      $\boldsymbol{E}_{t,d,p} = \max\left((1-\theta)\boldsymbol{E}_{t-1,d,p}, \left|\boldsymbol{r}^{\text{red}}_{d,p}\right|\right)$

      $\boldsymbol{\eta}_{t,d,p} = \gamma \min\left(\left(-\log(\boldsymbol{\beta}_{0,d,p}) \odot \boldsymbol{V}^{\odot -1}_{t,d,p}\right)^{\odot \frac{1}{2}}, \frac{1}{2}\boldsymbol{E}^{\odot -1}_{t,d,p}\right)$

      $\boldsymbol{R}_{t,d,p} = (1-\theta)\boldsymbol{R}_{t-1,d,p} + \boldsymbol{r}^{\text{red}}_{d,p} \odot \left(1 - \boldsymbol{\eta}_{t,d,p} \odot \boldsymbol{r}^{\text{red}}_{d,p}\right)/2 + \boldsymbol{E}_{t,d,p} \odot \mathbb{1}\{-2\boldsymbol{\eta}_{t,d,p} \odot \boldsymbol{r}^{\text{red}}_{d,p} > 1\}$

      $\boldsymbol{\beta}_{t,d,p} = K\boldsymbol{\beta}_{0,d,p} \odot \text{SoftMax}\left(-\boldsymbol{\eta}_{t,d,p} \odot \boldsymbol{R}_{t,d,p} + \log(\boldsymbol{\eta}_{t,d,p})\right)$

      $\boldsymbol{\beta}_{t,d,p} = \left(\mathcal{F}_{\phi} \circ \mathcal{H}_{\kappa} \circ \mathcal{S}_{\nu}\right)(\boldsymbol{\beta}_{t,d,p})$

    $\widetilde{\boldsymbol{X}}_{t,d} = \text{Sort}(\widetilde{\boldsymbol{X}}_{t,d})$

  **for** $k$ *in* $1, \dots, K$ **do**

    $\boldsymbol{w}_{t,k}(\boldsymbol{\mathcal{P}}) = \boldsymbol{\mathcal{H}}^{\text{mv}} \boldsymbol{B}^{\text{mv}} \boldsymbol{\beta}_{t,k} \boldsymbol{B}^{\text{pr}\prime} \boldsymbol{\mathcal{H}}^{\text{pr}}$

**end**

solution computed by BOA. Another typical particular case is constant weights, where each expert receives a specific weight. This can be calculated by setting both basis matrices, $B^{\text{mv}}$, and $B^{\text{pr}}$, to the unity Vector of length $D$ and $P$, respectively. This leads to weights without variation across marginals and probabilities (**Constant**). Setting only one of the basis matrices to the unity vector will result in weights that are constant over either marginals (**Constant Mv**) or probabilities (**Constant Pr**). Additionally, setting all smoothing matrices to identity produces pointwise weights, and optimizing $\lambda$ in the hat matrices concerning the predictive CRPS results in possibly smoothed weights. These cases are illustrated in Figures 2a-2d. The extensions discussed in Subsections 3.1 and 3.3 require the specification of various hyperparameters. There are many possible hyperparameters to choose from, and we do not have

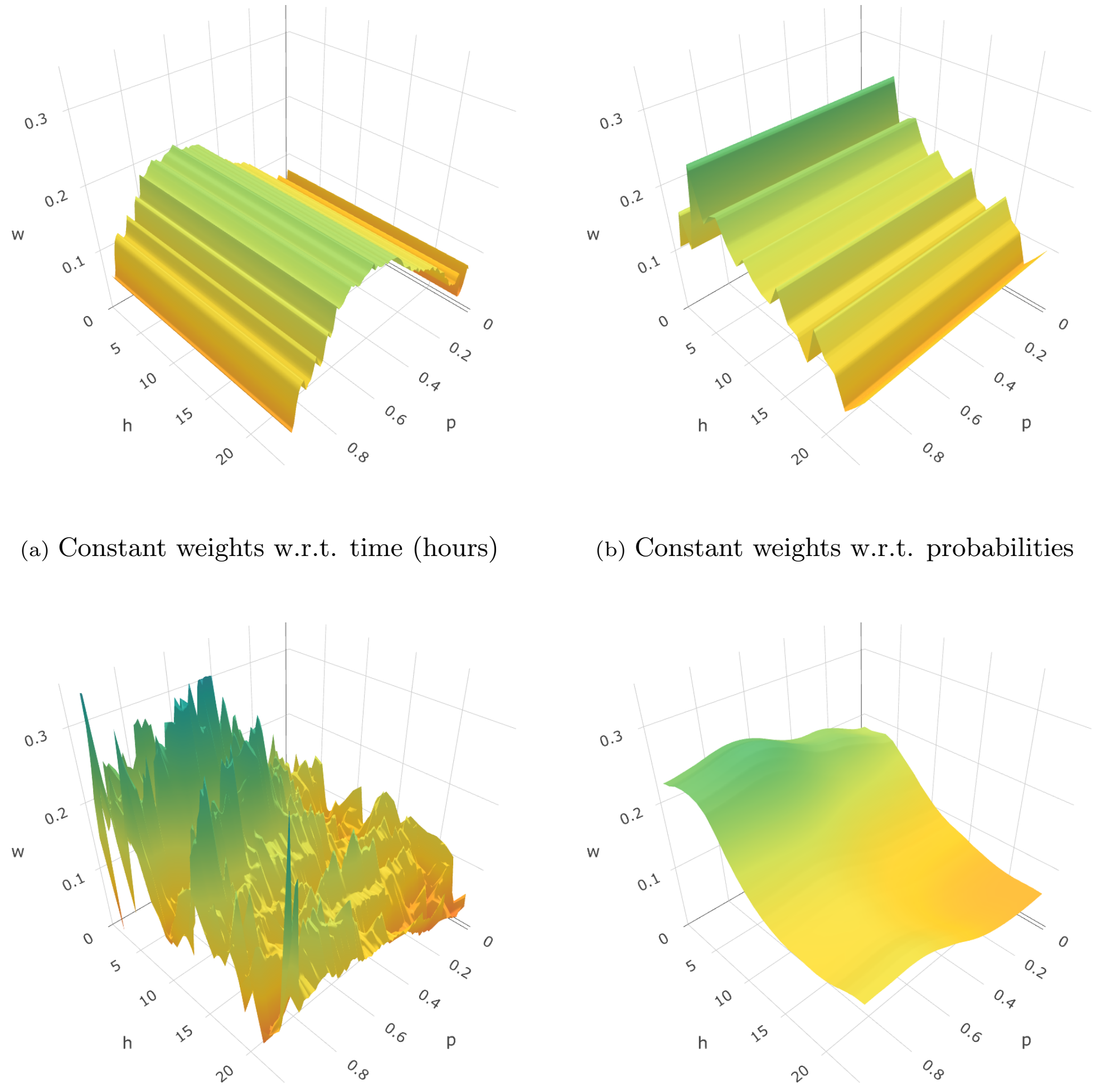

(a) Constant weights w.r.t. time (hours)

(b) Constant weights w.r.t. probabilities

(c) Optimized pointwise weights

(d) Optimized smoothed weights

Figure 2: Most recent weighs of JSU1 calculated using different specifications of Algorithm 2

any prior information on the best values. This means that it is impossible to test all combinations of these parameters in each iteration of the forecasting task. The latter would be ideal, but it is impractical due to the required computational resources. As a result, we need to use other, less demanding methods for tuning these hyperparameters. In this paper, we will utilize three approaches for tuning.

The first approach to hyperparameter tuning is using a sophisticated search algorithm based on random forest and optimizing towards the lowest CRPS on a subset of our observations (i.e., a training set). We utilize the R-Package `mlrMBO` to execute this optimization (Bischl et al., 2017). This approach brings one significant advantage: the search algorithm can efficiently search the considered space by repeatedly reevaluating the objective function. However, once the final set of hyperparameters is selected, it will remain constant throughout the forecasting task. Additionally, the tuning is only executed using a small subset of the dataset. This could be a problem as the chosen set of parameters may not be optimal for the rest of the dataset, especially if there are structural breaks. Hereinafter, we will refer to this approach as **Bayesian fix** as it fixes the hyperparameters after utilizing a Bayesian search algorithm.

The second approach uses the `online` function, which is included in the *profoc* R-Package (Berrisch & Ziel, 2023). It implements the proposed algorithm and an online tuning strategy for the hyperparameters. This strategy considers a random sample of all possible hyperparameter sets, and for each iteration, the combination with the lowest aggregate CRPS is chosen. In contrast to the **Bayesian fix** approach, we define all possible parameter sets before the learning task. However, this method dynamically selects the parameter set based on past performance, allowing for dynamic adjustments if underlying properties change. The most significant drawback of this approach is that only a random sample of the hyperparameter space is considered. However, this approach has the advantage of adjusting dynamically to changes in the data. Therefore, we will refer to this approach as **Sampling Online**.

It is also possible to combine both approaches. In this case, `mlrMBO` optimizes on a subset of the data. Afterward, *online* uses the parameter combinations that got proposed in the `mlrMBO` optimization. This has the potential to profit from efficient exploration of the hyperparameter space and the ability to adjust to changes in the data dynamically. After this, we will refer to this approach as **Bayesian Online**.

## 5. Application to Multivariate Probabilistic Day-Ahead Power Prices

In day-ahead electricity price forecasting, we consider the price $Y_{t,h}$ at day $t$ and product $h = 1, \dots, H$ of the day. For hourly electricity prices, we have $H = 24$, and therefore $h$ is often referred to as *hour*, see Ziel & Weron (2018). We consider the forecasts of Marcjasz et al. (2023), which covers the period from December 27, 2018 to December 31, 2020. These forecasts are based on German electricity market data starting in January 2015. Barunik & Hanus (2023) also use that data in their probabilistic forecasting study with the same design. They are hourly forecasts of eight models, i.e., neural network specifications. The forecasts are given as distributional parameters for each hour (i.e., $\boldsymbol{\mathcal{D}} = (1, \dots, 24)$) of all 736 Days. We use those distributional forecasts for calculating quantiles on the equidistant grid of percentiles

$\boldsymbol{\mathcal{P}} = (0.01, \ldots, 0.99)$.

The performance of combinations is mainly determined by two factors: the performance of the considered experts and the diversity between them. The first should naturally be high as an expert can only be beneficial if it provides valuable information; the latter is equally important since there is close to no benefit in combining very similar forecasts. Figure 3 shows the correlation between the experts. We show Pearson's correlation on the lower triangle, which takes values in $[-1, 1]$. In the upper triangle, we show the distance correlation. The distance correlation is a non-linear dependency measure that takes values in $[0, 1]$ and characterizes stochastic independence Székely et al. (2007). Unsurprisingly, we observe positive values for both dependence measures as all time series forecast the same target. However, all values are clearly below 1. This indicates diversity between experts, which is beneficial for the combination task.

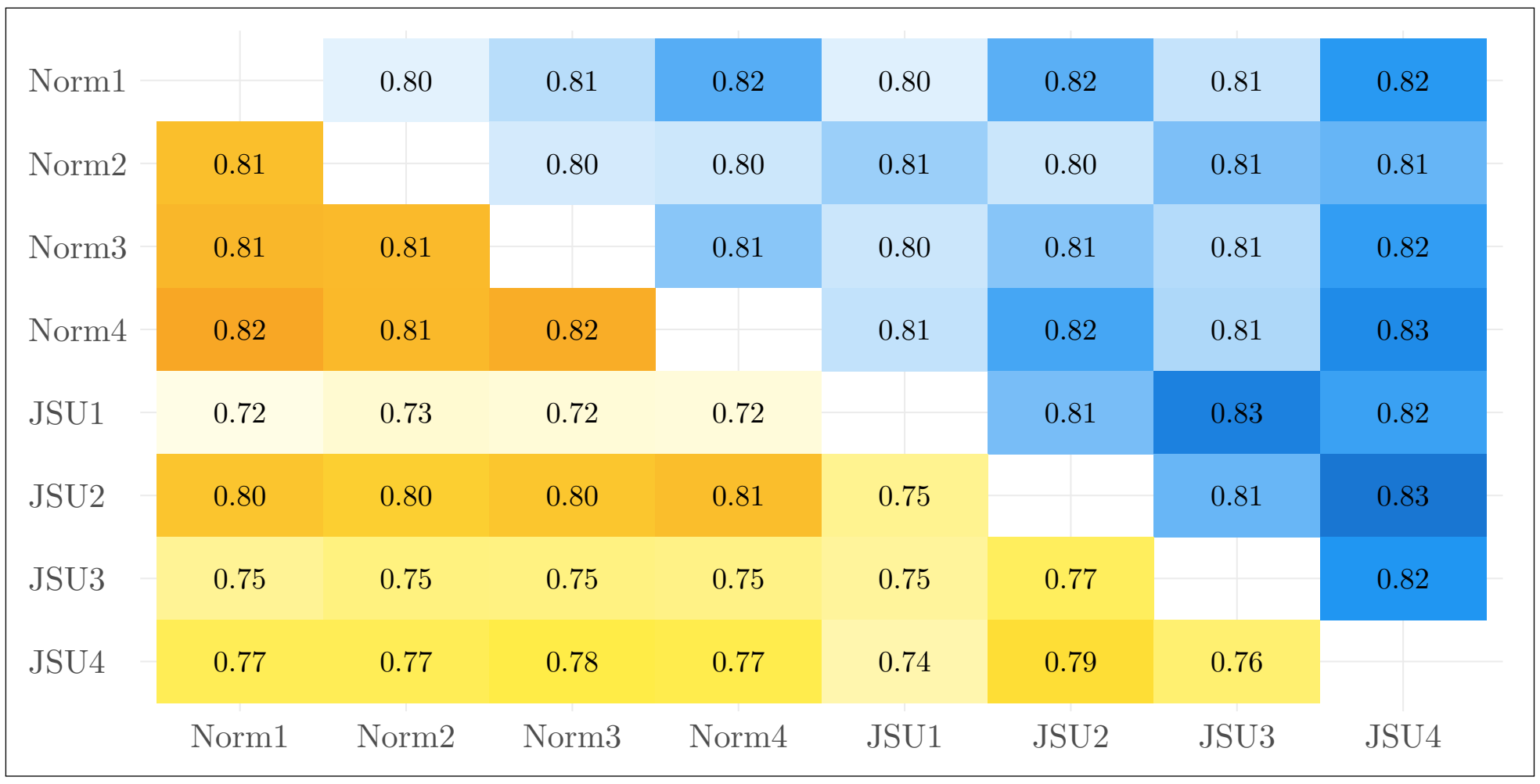

Figure 3: Correlation plot with Pearson's correlation on the lower triangle and distance correlation on the upper triangle.

The simulations of Berrisch & Ziel (2021) show superior performance for penalized smoothing compared to the basis smoothing approach. Therefore, we solely use the penalized smoothing approach for our learning task. We use 99 knots, i.e., one on each quantile.

We consider the knot placement and the other extensions discussed in Subsections 3.1 and 3.3. Table 1 summarizes the considered hyperparameters. That is, we have a total of 15 tuning parameters to optimize.

We conduct the forecasting task using the three tuning strategies **Bayesian fix**,

| Description | Notation | Range | Trafo | Model Specification | | | |
|---|---|---|---|---|---|---|---|
| | | | | Full | Smooth.Forget | Smooth | Forget |
| Forget Regret | $\theta$ | $-12, \ldots, 2$ | $2^x$ | ✓ | ✓ | | ✓ |
| Fixed Share | $\phi$ | $-15, \ldots, 0$ | $2^x$ | ✓ | | | |
| Soft Threshold | $\nu$ | $-15, \ldots, 0$ | $2^x$ | ✓ | | | |
| Hard Threshold | $\kappa$ | $-15, \ldots, 0$ | $2^x$ | ✓ | | | |
| Learning rate adjustment | $\gamma$ | $-1, \ldots, 1$ | $2^x$ | ✓ | | | |
| Penalized Smoothing Prob. | $\lambda^{\text{pr}}$ | $-5, \ldots, 15$ | $2^x$ | ✓ | ✓ | ✓ | |
| Penalized Smoothing Mult. | $\lambda^{\text{mv}}$ | $-5, \ldots, 15$ | $2^x$ | ✓ | ✓ | ✓ | |
| Knot Placement Prob. | $\mu^{\text{pr}}$ | $-1, \ldots, 1$ | $x^3/2.1 + 0.5$ | ✓ | | | |
| Knot Placement Prob. | $\sigma^{\text{pr}}$ | $-1, \ldots, 1$ | $x^3/1.1 + 1$ | ✓ | | | |
| Knot Placement Prob. | $c^{\text{pr}}$ | $-3, \ldots, 3$ | $x^3$ | ✓ | | | |
| Knot Placement Prob. | $\tau^{\text{pr}}$ | $-1, \ldots, 1$ | $x^3/1.1 + 1$ | ✓ | | | |
| Knot Placement Mult. | $\mu^{\text{mv}}$ | $-1, \ldots, 1$ | $x^3/2.1 + 0.5$ | ✓ | | | |
| Knot Placement Mult. | $\sigma^{\text{mv}}$ | $-1, \ldots, 1$ | $x^3/1.1 + 1$ | ✓ | | | |
| Knot Placement Mult. | $c^{\text{mv}}$ | $-3, \ldots, 3$ | $x^3$ | ✓ | | | |
| Knot Placement Mult. | $\tau^{\text{mv}}$ | $-1, \ldots, 1$ | $x^3/1.1 + 1$ | ✓ | | | |

Table 1: Considered hyperparameters with their ranges and respective transformation functions and three nested specifications (nested in the *Full* specification, which considers all of the above hyperparameters).

**Sampling Online**, and **Bayesian Online** discussed in Section 4. Marcjasz et al. (2023) used about half a year of data, i.e., the first (182) observations, as a burn-in period for hyperparameters to stabilize. With **Bayesian fix**, we use these first 182 observations. However, we do not evaluate the forecasts of the first 50 observations due to the elevated estimation uncertainty early in the learning process. We utilize the Krigin learner of `mlrMBO` to propose eight new points until the budget of 1000 points is exhausted. This is done in parallel. Then, the best hyperparameter set is used to conduct the forecast combination task with all 736 observations. For our final evaluation, we follow Marcjasz et al. (2023) again by excluding the first 182 observations. Appendix B.2 presents a detailed overview of the computation times on our Intel i5-12600K CPU.

For **sampling online**, we first divide the range of each hyperparameter into 16 equidistant values, apply the transformation function (see Table 1), and then randomly sample up to 2500 points from the resulting multivariate hyperparameter space. The online optimization process is then carried out as described in Section 4. As with **Bayesian fix**, the first 182 observations are excluded from the evaluation.

For **Bayesian Online**, we run **Bayesian fix** analogous to the above but with a reduced budget of 750 points to propose. These points are fed into the **Sampling Online** optimization.

In addition to tuning all 15 hyperparameters (**Full**), we examine three subsets of these hyperparameters. The first subset only includes penalized smoothing and forget (**Smooth.Forget**), the second subset only includes penalized smoothing (**Smooth**), and the last subset only includes forget (**Forget**). Note that the time required for computing **Smooth Forget** is reduced when using **Sampling Online** as the number of possible parameter combinations does not exceed 2500. The specifications are summarized in Table 1. We also report the performance of the **naive**, the performance of each expert, and the four special cases shown in Figure 2.

Table 2 summarizes the results. It reports the CRPS of each expert and the **naive** combination in the top row and the performance of different specifications of Algorithm 2. The combination schemes in Table 2 consider the full set of experts. Appendix B.1 presents the results for considering the Gaussian and JSU experts separately.

We also report the performance of **ML-Poly** and **EWA** weighting schemes as they are popular in the forecast combination literature (Gaillard et al., 2014; Jore et al., 2010; Dalalyan et al., 2012; Opschoor et al., 2017). V'yugin & Trunov (2022) and Zamo et al. (2021) use **EWA** together with the CRPS to receive constant weights across the whole distribution. This corresponds to the **B Constant Pr** scheme using **EWA**. The ML-Poly algorithm with the CRPS is used in Thorey et al. (2018); this corresponds to the **B Constant Pr** scheme using **ML-Poly**. However, **ML-Poly** and **EWA** have inferior convergence properties, compared to **BOA** (Berrisch & Ziel, 2021). That is, we do expect them to perform worse. As an additional benchmark, we report the performance of the Follow-The-Leader (FTL) strategy. This strategy selects the expert who had the smallest loss in the previous iteration (Huang et al., 2017). We always apply the gradient trick (see. Section 3).

We tested the hypothesis of equal accuracy in forecast performance between the **naive** model and the more sophisticated forecast combinations using the Diebold Mariano (DM) test (Diebold & Mariano, 2002). We apply this DM test with the small sample adjustment of Harvey et al. (1997, eq. (9)). Thereby we use the following loss differential: $\Delta_t^{\text{naive},x} = \|\boldsymbol{L}_t^{\text{naive}}\|_1 - \|\boldsymbol{L}_t^x\|_1$, where $\boldsymbol{L}_t^{\cdot}$ denotes the 24-dimensional vector of CRPS losses on day $t$ for the respective model and $\|\boldsymbol{L}\|_1$ denotes the $L_1$ norm of the former. The table cells are colored according to the resulting test statistic of

| JSU1 | JSU2 | JSU3 | JSU4 | Norm1 | Norm2 | Norm3 | Norm4 | Naive |
|---|---|---|---|---|---|---|---|---|
| 1.487 | 1.444 | 1.499 | 1.374 | 1.414 | 1.535 | 1.42 | 1.422 | **1.295** |

| Description | Parameter Tuning | BOA | ML-Poly | EWA |
|---|---|---|---|---|
| Constant | | 1.2933 | 1.2966 | 1.3188 |
| Pointwise | | 1.2936 | 1.3010 | 1.3101 |
| FTL | | 1.3752 | 1.3692 | 1.3863 |
| B Constant Pr | | 1.2936 | 1.3000 | 1.3432 |
| B Constant Mv | | 1.2918 | 1.2945 | 1.3076 |
| Forget | Bayesian Fix | 1.2930 | 1.2956 | 1.3096 |
| Full | Bayesian Fix | 1.2905 | 1.2902 | 1.2870 . |
| Smooth.forget | Bayesian Fix | 1.2911 | 1.2912 | 1.2869 . |
| Smooth | Bayesian Fix | 1.2918 | 1.2917 | 1.2873 . |
| Forget | Bayesian Online | 1.2855 ** | 1.2961 | 1.3098 |
| Full | Bayesian Online | 1.2919 | 1.2873 . | 1.2873 . |
| Smooth.forget | Bayesian Online | **1.2845** ** | 1.2862 * | **1.2864** . |
| Smooth | Bayesian Online | 1.2918 | 1.2918 | 1.2874 . |
| Forget | Sampling Online | 1.2855 ** | 1.2961 | 1.3114 |
| Full | Sampling Online | 1.2886 | **1.2861** * | 1.2873 . |
| Smooth.forget | Sampling Online | 1.2845 *** | 1.2867 * | 1.2866 . |
| Smooth | Sampling Online | 1.2918 | 1.2917 | 1.2877 . |

Coloring w.r.t. test statistic: <-5 -4 -3 -2 -1 0 1 2 3 4 >5

Table 2: CRPS scores (lower = better) of the individual experts and the *naive* combination (top), and different specifications (see Table 1) of Algorithm 2 (bottom). The cells are colored according to the test statistic of the DM-Test comparing the model in question to *Naive* using (greener means lower test statistic, i.e., better performance compared to *naive*). Significance is denoted as follows: .p≤0.1; *p≤0.05; **p≤0.01; ***p≤0.001.

the Diebold Mariano test.

We see that all individual experts perform worse compared to **naive**. The best results were obtained by the **Bayesian Online** approach and using equidistant knots, penalized smoothing, and a forget rate (**Smooth.Forget**). This solution yields a significant improvement over the naive combination. Comparing this solution with the smaller models **Smooth** and **Forget**, we conclude that forgetting contributes to most of the observed improvement. The importance of the forgetting factor indicates structural changes in the data. Further evidence comes from the fact that the dynamic **Bayesian Online** optimization generally outperforms parameter optimization using **Bayesian fix**. We did analyze the performance of the combination schemes on subsets of the data. However, the patterns are very similar to the ones observed in Table 2. Therefore, we do not report them here.

Further, BOA performs best compared to the other considered weighting schemes ML-Poly and EWA. Overall, forgetting and smoothing play a crucial role in the performance of the combination. Finally, regarding the hyperparameter tuning, we conclude that the dynamic optimization **Bayesian Online** and **Sampling Online** should be preferred to the static Bayesian optimization **Bayesian Fix**.

We also analyzed the issue of quantile crossing for all considered combination schemes. Quantile Crossing happened in at least one marginal on 67 of the 554 test-set days for the best performing scheme **Bayesian Online Smooth.Forget** using **BOA**. For brevity, we transferred the detailed discussion to Appendix B.3.

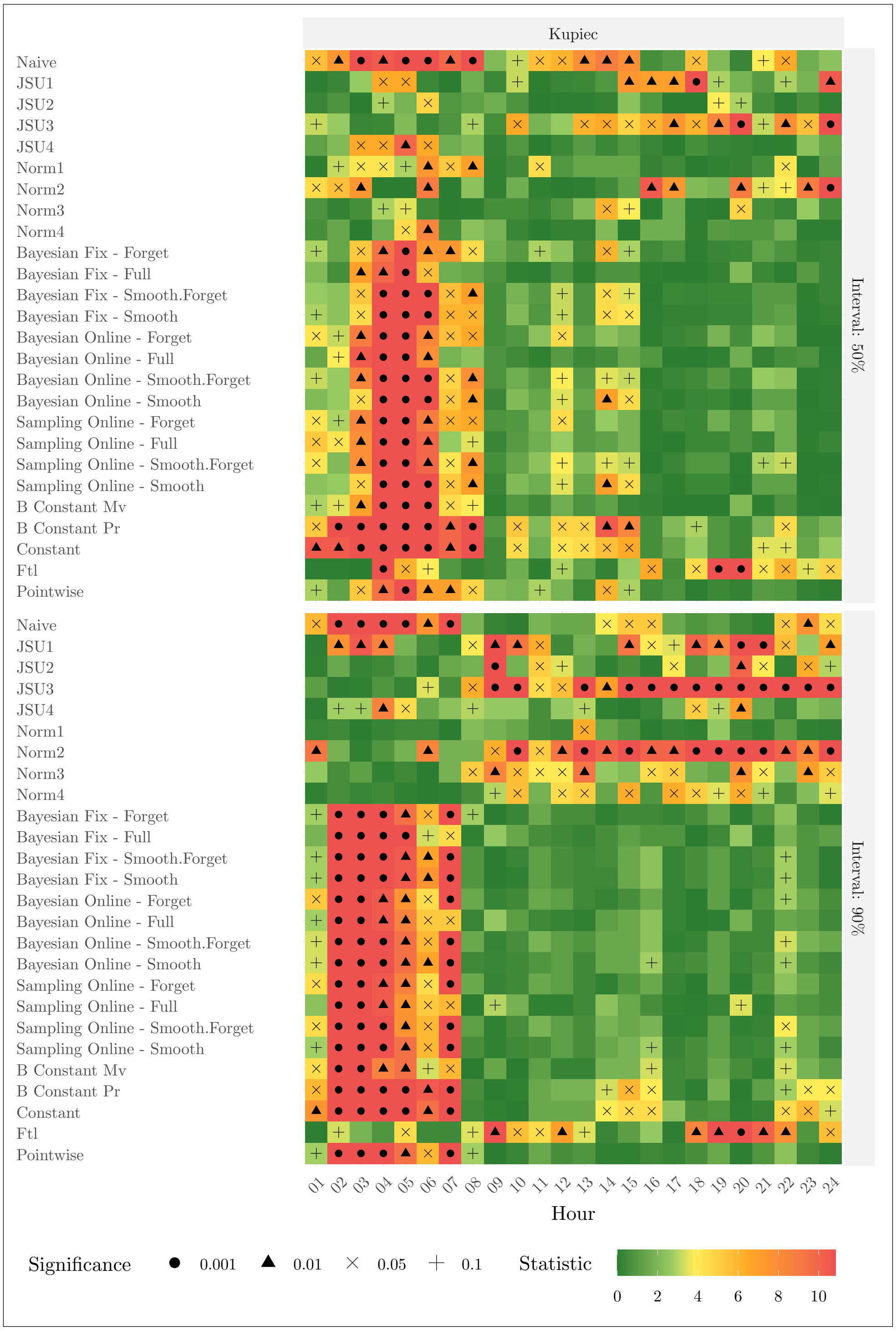

Figure 4: Significance and test statistics of the unconditional coverage test of Kupiec for the full set of experts (Kupiec et al., 1995). This table corresponds to the first row, and the **BOA** column of Table 2. The facets present 50% (top) and 90% (bottom) intervals, the symbols indicate significance, and the cells are colored w.r.t. the test statistics. Thereby, the upper limit (dark-red) corresponds to the 0.001 significance level.

We performed the Kupiec and Christoffersen tests for coverage (Kupiec et al., 1995; Christoffersen, 1998). Both tests are based on prediction interval violations. The Null hypothesis states that $\alpha\%$ of the observations lie outside the $100 - \alpha\%$ prediction interval. However, the Kupiec test ignores the potential autocorrelation of these violations. The Christoffersen test tests jointly for unconditional coverage, conditional coverage, and the temporal independence of the violations. However, the test solely considers temporal independence for the first time lag. Figure 4 presents the results of the Kupiec test. As in (Marcjasz et al., 2023), we must reject the Null at the 5% level for selected hours. Interestingly, the coverage is worse during the night hours. However, the coverage does not differ much between the combinations. The results of the Christoffersen test are attached in Appendix B.4. The results draw a more negative picture in general. In addition to the night hours, we observe multiple significant hours during the day. That is, the temporal independence of the violations seems to be present, particularly during the afternoon.

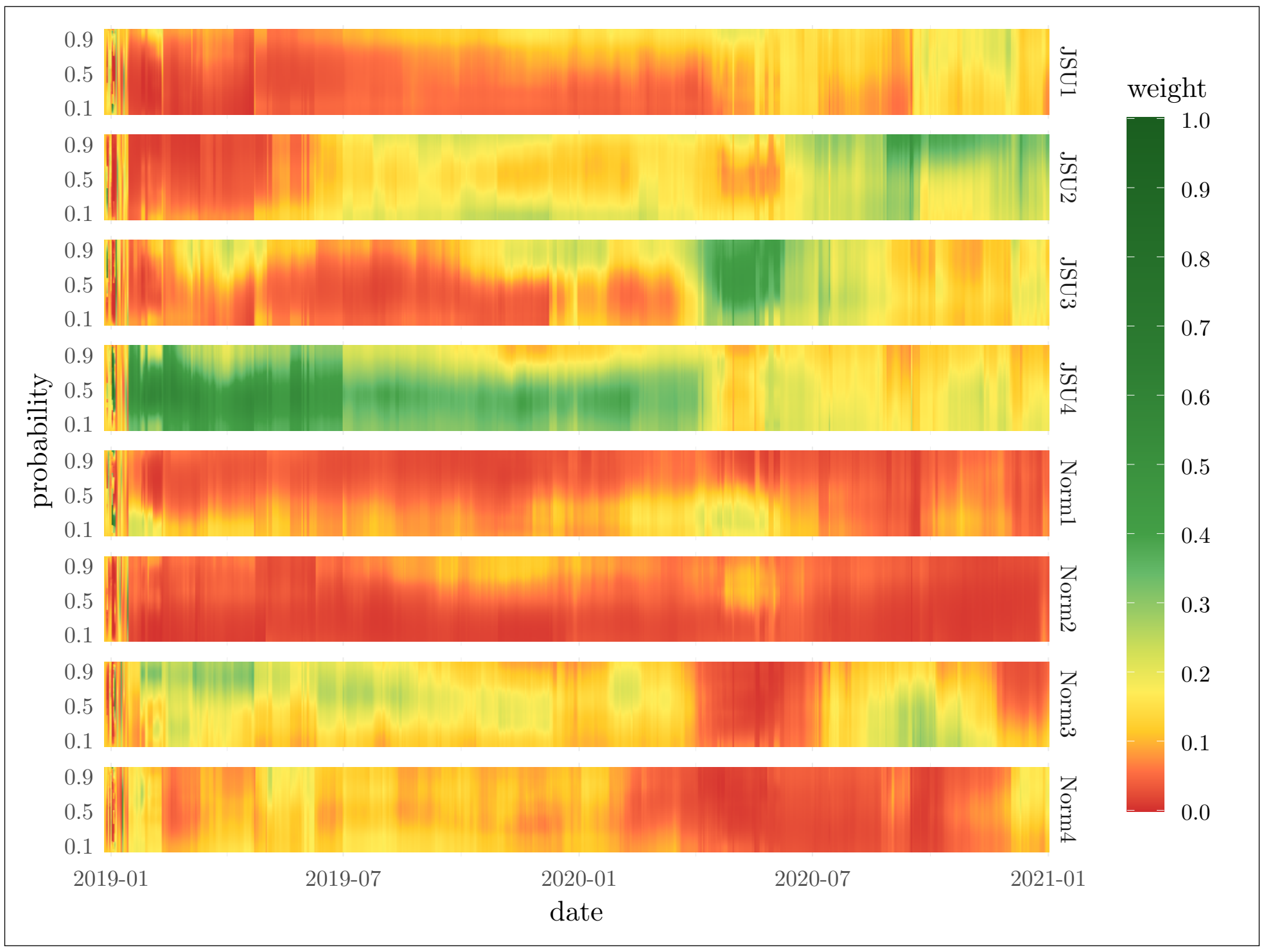

Figure 5: Temporal evolution of the weights of **Smooth.Forget Bayesian Online** at hour 16 across all 99 probabilities

Figures 5 and 6 provide a more detailed analysis of the proposed model (**Smooth.Forget**, **Bayesian Online**). They depict the temporal evolution of the weights for each ex-pert. Thereby, Figure 5 presents the temporal evolution across probabilities for hour 16 of the day. After a brief initial burn-in period, the weights of the eight experts stabilized. There is a higher degree of variability in the center of the distribution. The weights are close to the uniform solution at the tails, with only a few exceptions. JSU4 strongly influences the combined value in the center of the distribution until around April 2020. After that point, the weight of JSU4 decreases, and JSU3 becomes more prominent. Further, there seem to be noticeable changes in the weights around June 2019 and April 2020, suggesting possible structural changes. These structural changes potentially lead to the dynamic hyperparameter tuning **Sampling Online** and **Bayesian Online** performing better than **Bayesian fix** due to the ability of the hyperparameters to adapt the changing data.

Figure 6 shows the temporal evolution of the weights at the median across all 24 hours. However, the high weights for JSU4 (see Figure 5) are only present in the afternoon and evening. Additionally, the plot reveals structural changes around June 2019 (regarding JSU4 and NORM4) and March 2020 (concerning JSU3, JSU4, NORM3, and NORM4). The latter coincides with the German government's introduction of strict COVID-19 measures. So, we suspect that these changes are due to changes in the power market due to the adjusted behavior of the market participants.

Both graphs indicate that the weights vary with time, hours, and quantiles. Thus, a flexible approach like the one proposed in this study seems appropriate. Lastly, the weights show less smoothing across hours than across quantiles.

Figure 7 presents the parameters used by the proposed *online.sm.fr* specification. This approach optimizes the forgetting rate and the two smoothing penalties. All parameters need some time to stabilize. However, the chosen burn-in period (marked in grey) seems to suffice for the most part. Further, we observe an increasing forgetting as the learning progresses and a consistent smoothing level across both dimensions. Lastly, the weights are getting more smoothed across probabilities than across hours (see also Figure 2d, which presents the most recent weights of this solution across hours and probabilities). Note, however, that the parameters show more persistence as time progresses. This is because hyperparameters are selected based on the cumulative past performance. For larger time series, it is, therefore, advisable to introduce a forgetting factor to the cumulative past performance to ensure reasonably fast adjustment of the hyperparameters to structural changes in the data. Our implementation includes this setting. However, we did not use it in this paper as the time series is relatively

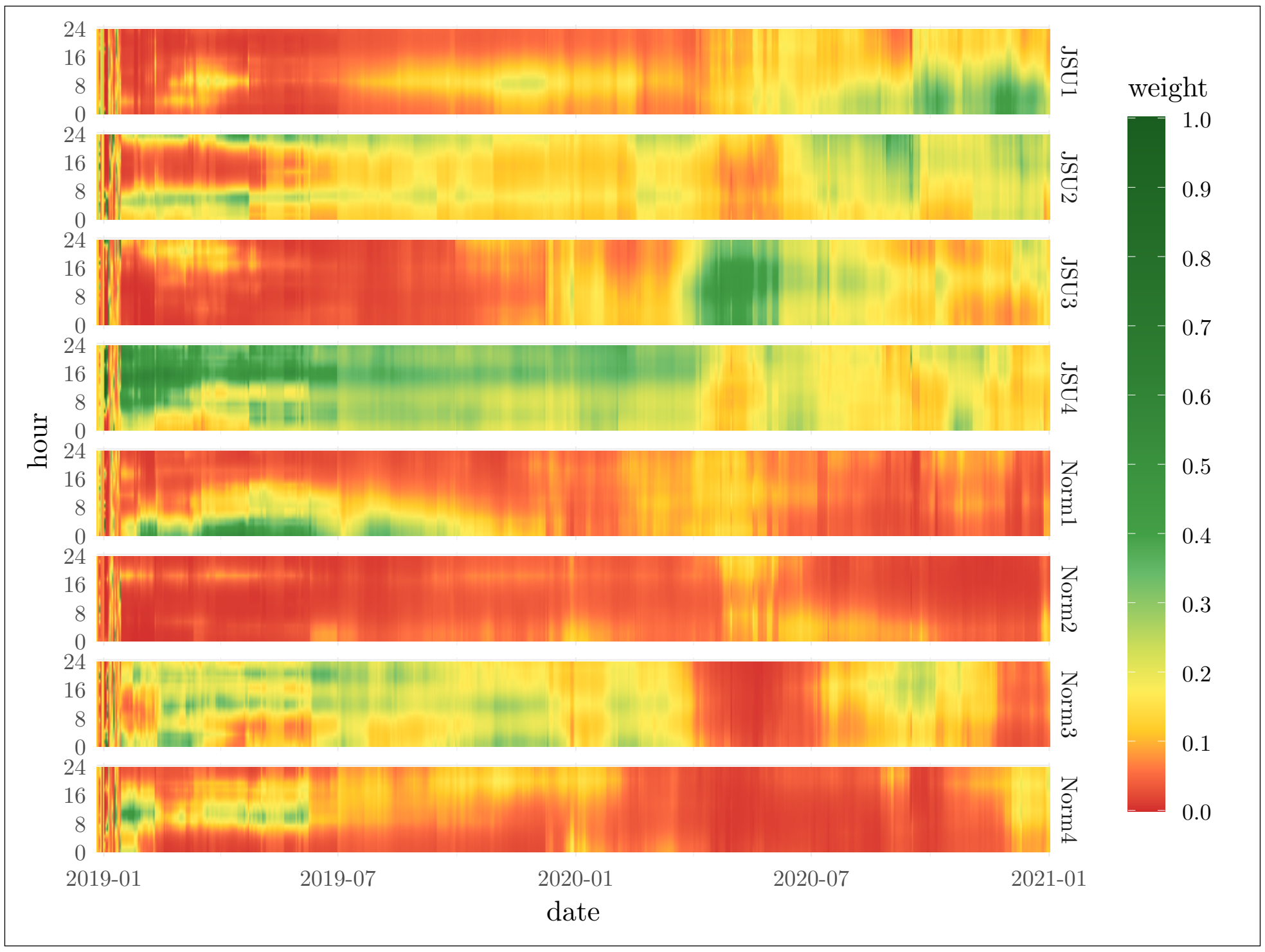

Figure 6: Temporal evolution of the weights of **Smooth.Forget Bayesian Online** at the median (50% quantile) across all 24 hours

short.

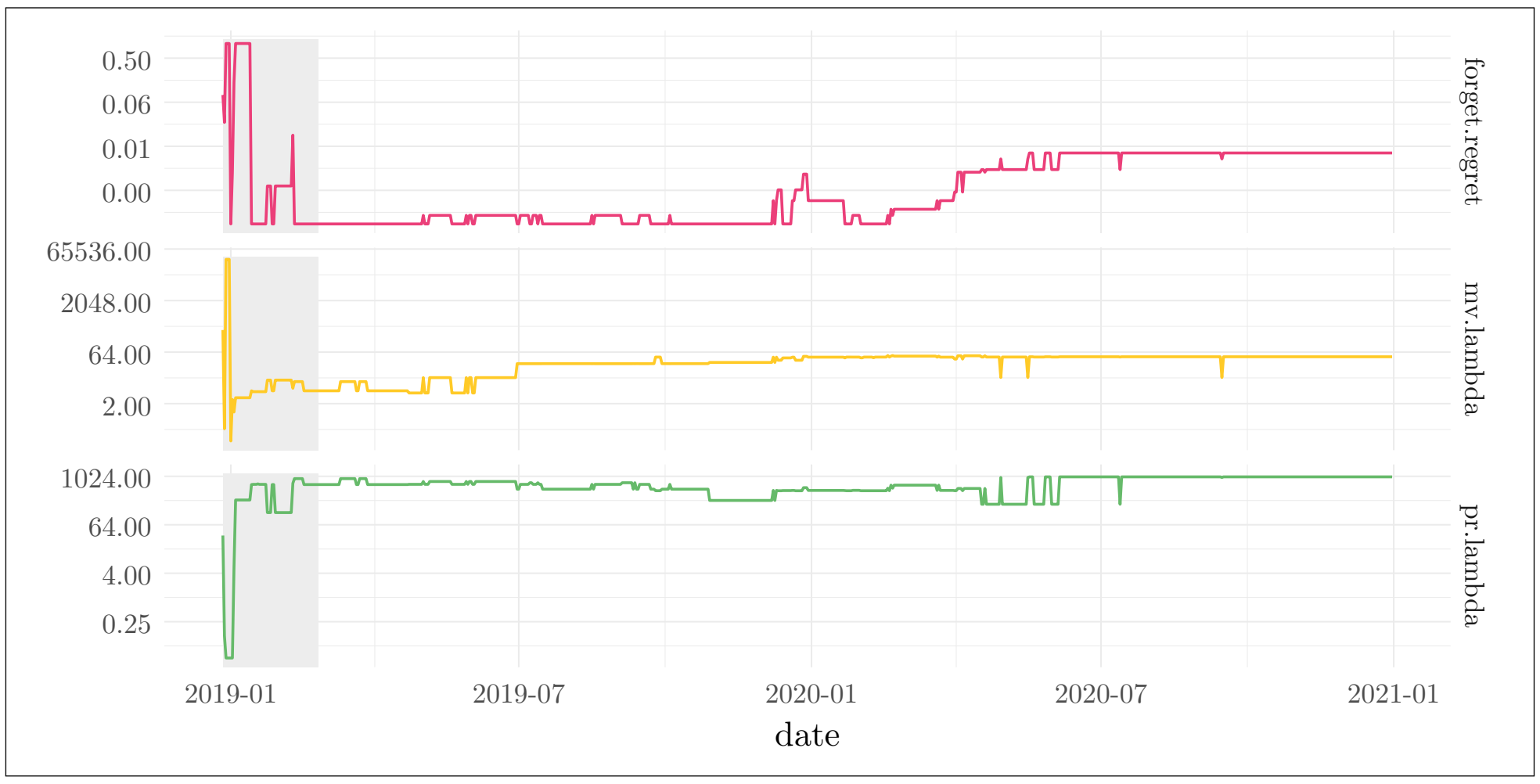

Figure 7: Hyperparameter values of **Smooth.Forget Bayesian Online**. The grey area represents the burn-in period which is not used for evaluation.

## 6. Conclusion

This paper proposes a novel method for combining multivariate probabilistic forecasts, considering dependencies between quantiles and marginals through a smoothing procedure. The first discussed smoothing method reduces the dimensionality of the regret using Basis matrices. It bridges the gap between pointwise and constant weight optimization. The second method involves smoothing the weights by penalized smoothing. The proposed method extends the (online) *CRPS learning* algorithm and is based on gradient-based EWA, which yields fast convergence rates. We also discuss possible extensions to the algorithm, such as forgetting and shrinkage operators. We also discuss how non-equidistant knots can be used in penalized smoothing, i.e., how the penalty term needs to be adjusted.

We apply the proposed methodology to multivariate probabilistic forecasts for day-ahead electricity prices. For hyperparameter tuning of Algorithm 2, we compare three strategies **Bayesian fix** and **Sampling Online** and a mixture of both **Bayesian Online**.

The best performance is obtained by optimizing **Bayesian Online** with penalized smoothing and a forget rate. This specification significantly improves over the *naive* combination. The forget rate contributes to most of the improvement, indicating structural changes in the data. The combination weights stabilize quickly after a short burn-in period. However, we observed some structural change in the weights around November 2019, which remains puzzling. The BOA weighting scheme obtained the best results out of the three considered online learning algorithms.

The smoothing methods used in this paper assume a simple metric structure. This assumption is reasonable for time series that are combined over the forecast horizon, as the order is naturally determined by time. In other cases, the adjacency structure may be more complex. For example, in spatial data, the first and last coordinates are often adjacent (e.g., forecasting a meteorological feature along the equator). Other spatial structures (e.g., across the globe) may be even more complex. We can account for these structures using the Laplacian matrix (graph Laplacian, admittance matrix, or Kirchhoff matrix) as the smoothing penalty. The Laplacian can be calculated from the so-called adjacency matrix, which describes the metric structure of the data. This procedure generalizes our methodology to more complex adjacency structures. This is a promising topic for future research.

The approach proposed in this paper is based on the CRPS as it combines the marginals of a multivariate distribution. This means that the experts only need to report the marginals. The copula is not needed and remains untouched. A recent study confirms that CRPS learning leads to a significant improvement in terms of CRPS. However, this improvement did not translate to its full extent into increased profits in a trading strategy (Nitka & Weron, 2023). An obvious question regarding multivariate forecast combination is if and how the copula can be incorporated into the CRPS learning framework. The Energy Score (ES) would be one option to base the weight updates on. However, a decomposition similar to 2 does not exist. Therefore, the ES is suitable only for describing the performance of a joint distribution as a whole. We can use the ES to derive constant combination weights across the entire distribution. However, this will ignore any variation in the experts' performance across the joint distribution (e.g., across quantiles). Limited research exists on so-called $\alpha$-projection quantiles (Hallin et al., 2010; Kong & Mizera, 2012). Meng et al. (2023) show a disintegration of the ES using $\alpha$-projection quantiles in their preprint. This may be a promising starting point for incorporating the copula into the flexible CRPS learning framework. For now, this remains a challenging topic for the future.

Overall, the empirical results approve the proposed combination algorithm, which uses adaptive weights across time, quantiles, and marginals. The algorithm surpasses the performance of simpler weighting schemes such as the *naive* in terms of the CRPS. A fast C++ implementation of the proposed algorithm is available in the open-source R-Package *profoc* (Berrisch & Ziel, 2023).

## 7. Data and Code Availability

The data and code used to produce Section 5 is available on github.com/BerriJ/mcrps. The repository contains detailed instructions for replicating the results of this paper.

## Appendix A. Initialization of Algorithm 2

---

**Algorithm 3:** Initialization of Algorithm 2

---

**input:** 4-dimensional array of expert predictions $(\widehat{X})_{t,d,p,k}$ and matrix of prediction targets $\boldsymbol{Y}_{t,d}$ for $t = 1, \dots, T$, $d \in \boldsymbol{\mathcal{D}} = (d_1, \dots, d_D)$, $p \in \boldsymbol{\mathcal{P}} = (p_1, \dots, p_P)$, $k = 1, \dots, K$

**initialize:**

$\boldsymbol{w}_0 = 1/K$

$\boldsymbol{E}_0 = \boldsymbol{V}_0 = \boldsymbol{R}_0 = \boldsymbol{0}$

$\boldsymbol{\beta}_0 = \text{pinv}\left(\boldsymbol{B}^{\text{mv}}\right) \boldsymbol{w}_0 \left(\boldsymbol{\mathcal{P}}\right) \text{pinv}\left(\boldsymbol{B}^{\text{pr}}\right)'$

$\boldsymbol{\mathcal{H}}^{\text{pr}} = \boldsymbol{B}^{\text{pr}}(\boldsymbol{B}^{\text{pr}\prime}\boldsymbol{B}^{\text{pr}} + \lambda^P(\alpha \boldsymbol{D}_1'\boldsymbol{D}_1 + (1-\alpha)\boldsymbol{D}_2'\boldsymbol{D}_2))^{-1}\boldsymbol{B}^{\text{pr}\prime}$

$\boldsymbol{\mathcal{H}}^{\text{mv}} = \boldsymbol{B}^{\text{mv}}(\boldsymbol{B}^{\text{mv}\prime}\boldsymbol{B}^{\text{mv}} + \lambda^D(\alpha \boldsymbol{D}_1'\boldsymbol{D}_1 + (1-\alpha)\boldsymbol{D}_2'\boldsymbol{D}_2))^{-1}\boldsymbol{B}^{\text{mv}\prime}$

---

## Appendix B. Supplementary Results

### *Appendix B.1. Expert Subsets*

| JSU1 | JSU2 | JSU3 | JSU4 | Norm1 | Norm2 | Norm3 | Norm4 | Naive |
|---|---|---|---|---|---|---|---|---|
| 1.487 | 1.444 | 1.499 | 1.374 | | | | | **1.299** |

| Description | Parameter Tuning | BOA | ML-Poly | EWA |
|---|---|---|---|---|
| Constant | | 1.2950 | 1.2968 | 1.3267 |
| Pointwise | | 1.2928 | 1.3041 | 1.3095 |
| FTL | | 1.3832 | 1.3607 | 1.3751 |
| B Constant Pr | | 1.2946 | 1.3016 | 1.3465 |
| B Constant Mv | | 1.2940 | 1.2966 | 1.3209 |
| Forget | Bayesian Fix | 1.2906 . | 1.3053 | 1.3136 |
| Full | Bayesian Fix | 1.2951 | 1.2912 . | 1.2942 |
| Smooth.forget | Bayesian Fix | 1.2913 . | 1.2902 . | 1.2927 |
| Smooth | Bayesian Fix | 1.2915 | 1.2938 | 1.2926 |
| Forget | Bayesian Online | 1.2909 * | 1.3059 | 1.3121 |
| Full | Bayesian Online | 1.2931 | 1.2909 . | 1.2936 |
| Smooth.forget | Bayesian Online | **1.2896** * | 1.2903 . | 1.2926 |
| Smooth | Bayesian Online | 1.2907 . | 1.2930 | 1.2925 |
| Forget | Sampling Online | 1.2907 * | 1.3054 | 1.3121 |
| Full | Sampling Online | 1.2978 | 1.2923 | 1.2945 |
| Smooth.forget | Sampling Online | 1.2902 * | **1.2901** . | **1.2920** |
| Smooth | Sampling Online | 1.2907 . | 1.2930 | 1.2921 |

Coloring w.r.t. test statistic: <-5 -4 -3 -2 -1 0 1 2 3 4 >5

Table B.3: CRPS scores (lower = better) of the JSU experts and the *naive* combination (top), and different specifications (see Table 1) of Algorithm 2 (bottom). The cells are colored according to the test statistic of the DM-Test comparing the model in question to *Naive* using (greener means lower test statistic, i.e., better performance compared to *naive*). Significance is denoted as follows: .p≤0.1; *p≤0.05; **p≤0.01; ***p≤0.001. The results correspond to the results of the *DDNN-JSU-qEns* row of Table 1 in Marcjasz, Narajewski, Weron & Ziel (2023)

| JSU1 | JSU2 | JSU3 | JSU4 | Norm1 | Norm2 | Norm3 | Norm4 | Naive |
|---|---|---|---|---|---|---|---|---|
| | | | | 1.414 | 1.535 | 1.42 | 1.422 | **1.348** |

| Description | Parameter Tuning | BOA | ML-Poly | EWA |
|---|---|---|---|---|
| Constant | | 1.3516 | 1.3541 | 1.3512 |
| Pointwise | | 1.3486 | 1.3549 | 1.3570 |
| FTL | | 1.4021 | 1.3972 | 1.4042 |
| B Constant Pr | | 1.3505 | 1.3543 | 1.3635 |
| B Constant Mv | | 1.3492 | 1.3516 | 1.3514 |
| Forget | Bayesian Fix | 1.3483 | 1.3543 | 1.3575 |
| Full | Bayesian Fix | 1.3525 | 1.3519 | 1.3483 |
| Smooth.forget | Bayesian Fix | 1.3480 | 1.3500 | 1.3485 |
| Smooth | Bayesian Fix | 1.3482 | 1.3501 | 1.3485 |
| Forget | Bayesian Online | 1.3396 *** | 1.3498 | 1.3530 |
| Full | Bayesian Online | 1.3441 | 1.3465 | 1.3435 |
| Smooth.forget | Bayesian Online | 1.3420 * | 1.3426 | 1.3452 |
| Smooth | Bayesian Online | 1.3473 | 1.3496 | 1.3481 |
| Forget | Sampling Online | 1.3395 *** | 1.3493 | 1.3517 |
| Full | Sampling Online | 1.3458 | 1.3417 . | 1.3428 |
| Smooth.forget | Sampling Online | **1.3390** *** | **1.3412** . | **1.3408** . |
| Smooth | Sampling Online | 1.3473 | 1.3495 | 1.3481 |

Coloring w.r.t. test statistic: <-5 -4 -3 -2 -1 0 1 2 3 4 >5

Table B.4: CRPS scores (lower = better) of the gaussian experts and the *naive* combination (top), and different specifications (see Table 1) of Algorithm 2 (bottom). The cells are colored according to the test statistic of the DM-Test comparing the model in question to *Naive* using (greener means lower test statistic, i.e., better performance compared to *naive*). The significance is denoted as follows: .p≤0.1; *p≤0.05; **p≤0.01; ***p≤0.001. The results correspond to the results of the *DDNN-N-qEns* row of Table 1 in Marcjasz, Narajewski, Weron & Ziel (2023)

*Appendix B.2. Computation Time*

Table B.5 reports the time needed to run the entire combination task, including the complete set of eight experts on our Intel i5-12600K CPU. However, note that the reported times concerning the **Bayesian Online** schemes do not include the Bayesian hyperparameter optimization but only the Online Optimization. That is caused by our implementation that uses the first 750 points of the respective **Bayesian Fix** optimization, which we run beforehand. Consequently, the reported times for **Bayesian Online** only account for roughly 25% of the total time needed to run the whole combination task. Computing the **naive** combination is significantly faster than running these more sophisticated combination schemes. The time needed is negligible.

| Description | Parameter Tuning | BOA | ML-Poly | EWA |
|---|---|---|---|---|
| Constant | | **0.0302** | 0.0309 | 0.0347 |
| Pointwise | | 0.0595 | 0.0548 | 0.0571 |
| FTL | | 0.0484 | 0.0391 | 0.0420 |
| B Constant Pr | | 0.0473 | 0.0455 | 0.0452 |
| B Constant Mv | | 0.0455 | **0.0307** | **0.0316** |
| Forget | Bayesian fix | 25.2240 | 26.0974 | 24.3612 |
| Full | Bayesian fix | 72.2383 | 70.3925 | 70.1709 |
| Smooth.forget | Bayesian fix | 46.1368 | 68.7141 | 39.6726 |
| Smooth | Bayesian fix | 32.8277 | 31.1237 | 32.0225 |
| Forget | Bayesian online | 11.5740 | 8.8163 | 8.5585 |
| Full | Bayesian online | 15.6165 | 12.5285 | 12.0932 |
| Smooth.forget | Bayesian online | 14.0724 | 11.4029 | 10.3729 |
| Smooth | Bayesian online | 14.3783 | 11.2084 | 10.7598 |
| Forget | Sampling online | 0.2614 | 0.1989 | 0.1934 |
| Full | Sampling online | 51.0216 | 41.8030 | 40.6413 |
| Smooth.forget | Sampling online | 44.8544 | 35.5308 | 33.7815 |
| Smooth | Sampling online | 4.6064 | 3.6358 | 3.6182 |

Table B.5: Computation times in minutes for all combination schemes considering the full set of experts. Colored w.r.t computation time (green = lower, red = higher).

*Appendix B.3. Quantile Crossing*

Table B.6 presents the prevalence of quantile crossing for all combination schemes. For conciseness, we aggregated over all marginals. The table reports the number of days on which quantile crossing happened in at least one of the 24 marginals. We see that quantile crossing is an issue for schemes without a smoothing component. Further, we see that smoothing is an effective tool for reducing quantile crossings. However, it only partially eliminates quantile crossings.

| Description | Parameter Tuning | BOA | ML-Poly | EWA |
|---|---|---|---|---|
| Constant | | 0 | 0 | 0 |
| Pointwise | | 554 | 554 | 554 |
| FTL | | 554 | 554 | 554 |
| B Constant Pr | | 0 | 0 | 0 |
| B Constant Mv | | 471 | 550 | 554 |
| Forget | Bayesian Fix | 554 | 554 | 554 |
| Full | Bayesian Fix | 8 | 0 | 0 |
| Smooth.forget | Bayesian Fix | 4 | 0 | 0 |
| Smooth | Bayesian Fix | 6 | 0 | 0 |
| Forget | Bayesian Online | 554 | 554 | 554 |
| Full | Bayesian Online | 101 | 259 | 0 |
| Smooth.forget | Bayesian Online | 67 | 6 | 0 |
| Smooth | Bayesian Online | 5 | 2 | 0 |
| Forget | Sampling Online | 554 | 554 | 554 |
| Full | Sampling Online | 1 | 26 | 167 |
| Smooth.forget | Sampling Online | 7 | 4 | 0 |
| Smooth | Sampling Online | 5 | 1 | 0 |

Table B.6: Number of days on which quantile crossing happened in at least one of the 24 marginals (considering the test period of 554 observations).

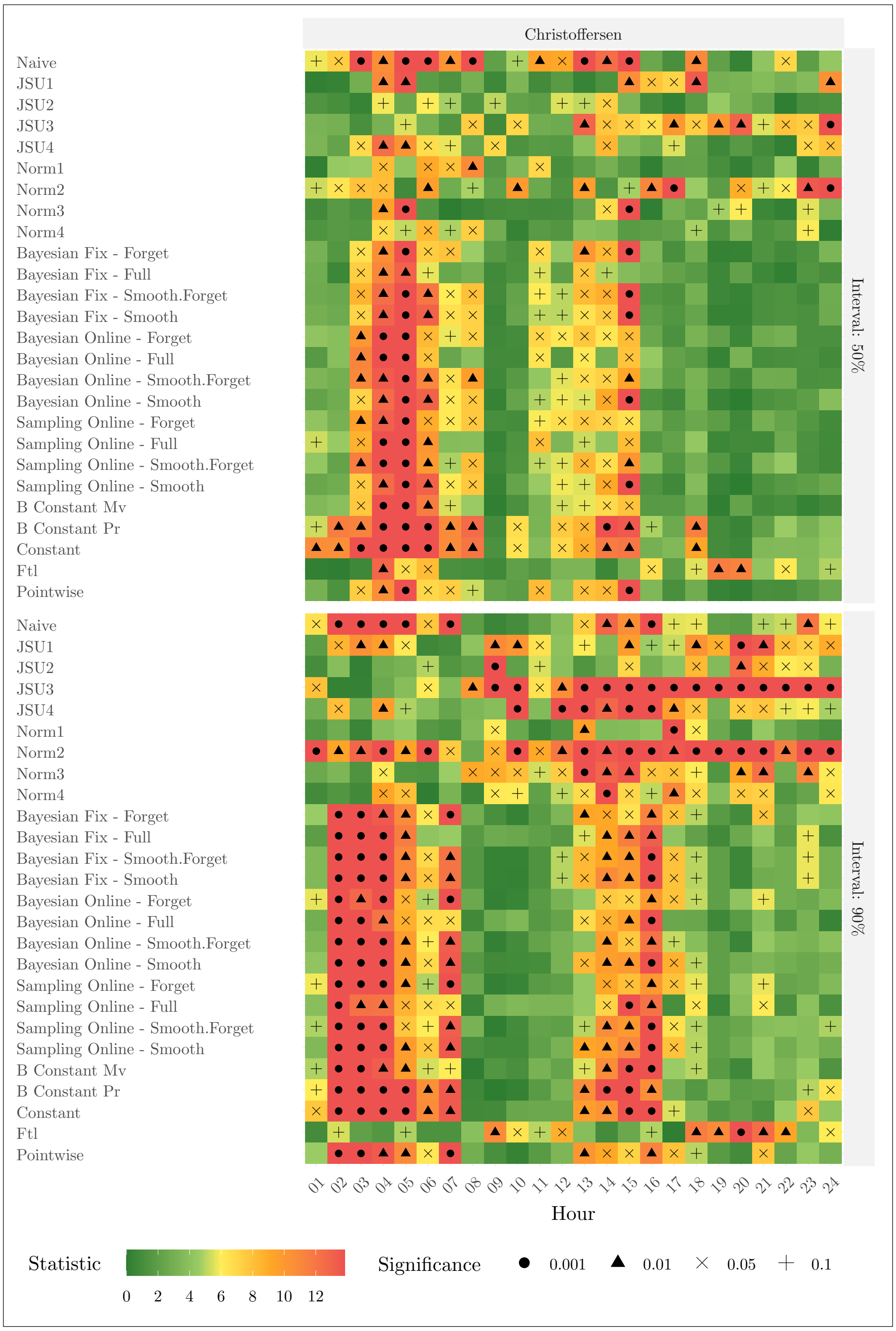

Figure B.8: Significance and test statistics of the Christoffersen coverage test for the full set of experts (Christoffersen, 1998). This table corresponds to the first row, and the **BOA** column of Table 2. The facets present 50% (top) and 90% (bottom) intervals, the symbols indicate significance, and the cells are colored w.r.t. the test statistics. Thereby, the upper limit (dark-red) corresponds to the 0.001 significance level

.